\pdfoutput=1

\documentclass[11pt]{article}
\DeclareUnicodeCharacter{FF0C}{,}

\usepackage[final]{acl}

\usepackage{times}
\usepackage{latexsym}

\usepackage[T1]{fontenc}

\usepackage[utf8]{inputenc}

\usepackage{microtype}

\usepackage{inconsolata}

\usepackage{graphicx}
\usepackage{booktabs}
\usepackage{multirow}
\usepackage{xcolor}
\usepackage{xspace}
\usepackage{colortbl}
\usepackage{tabulary}
\definecolor{lightgray}{gray}{0.95}
\definecolor{lightgray2}{gray}{0.9}
\usepackage{subcaption}
\usepackage{pgfplots}
\pgfplotsset{compat=1.18}
\usepackage{amsmath} 
\usepackage{pgfplotstable}
\usepackage{pgf-pie}   

\newcommand{\ie}{\emph{i.e.,}\xspace}
\newcommand{\eg}{\emph{e.g.,}\xspace}

\newcommand{\framework}{\textsc{FineReason}\xspace}
\title{\framework: Evaluating and Improving LLMs' Deliberate Reasoning through Reflective Puzzle Solving}

\author{
  Guizhen Chen$^{1,2,}$\thanks{Guizhen and Chaoqun are under the Joint PhD Program between Alibaba and NTU.}
  \quad Weiwen Xu$^{2,}$\thanks{Corresponding authors.}
  \quad Hao Zhang$^{2,}$\footnotemark[2]  
  \quad Hou Pong Chan$^{2}$\\
   \quad \textbf{Chaoqun Liu}$^{1,2,}$\footnotemark[1]
   \quad \textbf{Lidong Bing}$^{2}$
   \quad \textbf{Deli Zhao}$^{2,3}$
   \quad \textbf{Anh Tuan Luu}$^{1}$
   \quad \textbf{Yu Rong}$^{2,3}$ \\
  $^1$ Nanyang Technological University, Singapore \\  $^2$ DAMO Academy, Alibaba Group, Singapore \\ 
  $^3$ Hupan Lab, Hangzhou, China 
}

\begin{document}
\maketitle
\begin{abstract}
Many challenging reasoning tasks require not just rapid, intuitive responses, but a more deliberate, multi-step approach. Recent progress in large language models (LLMs) highlights an important shift from the  ``System 1'' way of quick reactions to the ``System 2'' style of reflection-and-correction problem solving. However, current benchmarks heavily rely on the final-answer accuracy, leaving much of a model's intermediate reasoning steps unexamined. This fails to assess the model's ability to reflect and rectify mistakes within the reasoning process. To bridge this gap, we introduce \framework, a logic-puzzle benchmark for systematic evaluation of LLMs' reasoning capabilities. Each puzzle can be decomposed into atomic steps, making it ideal for rigorous validation of intermediate correctness. Building on this, we introduce two tasks: \textbf{state checking} and \textbf{state transition}, for a comprehensive evaluation of how models assess the current situation and plan the next move. To support broader research, we also provide a puzzle training set aimed at enhancing general reasoning. We show that models trained on our state checking and transition data demonstrate gains in mathematical reasoning by up to $5.1\%$.\footnote{\url{https://github.com/DAMO-NLP-SG/FineReason}}

\end{abstract}

\section{Introduction}

\begin{figure}[t!]
 \centering
    \includegraphics[width=\linewidth]{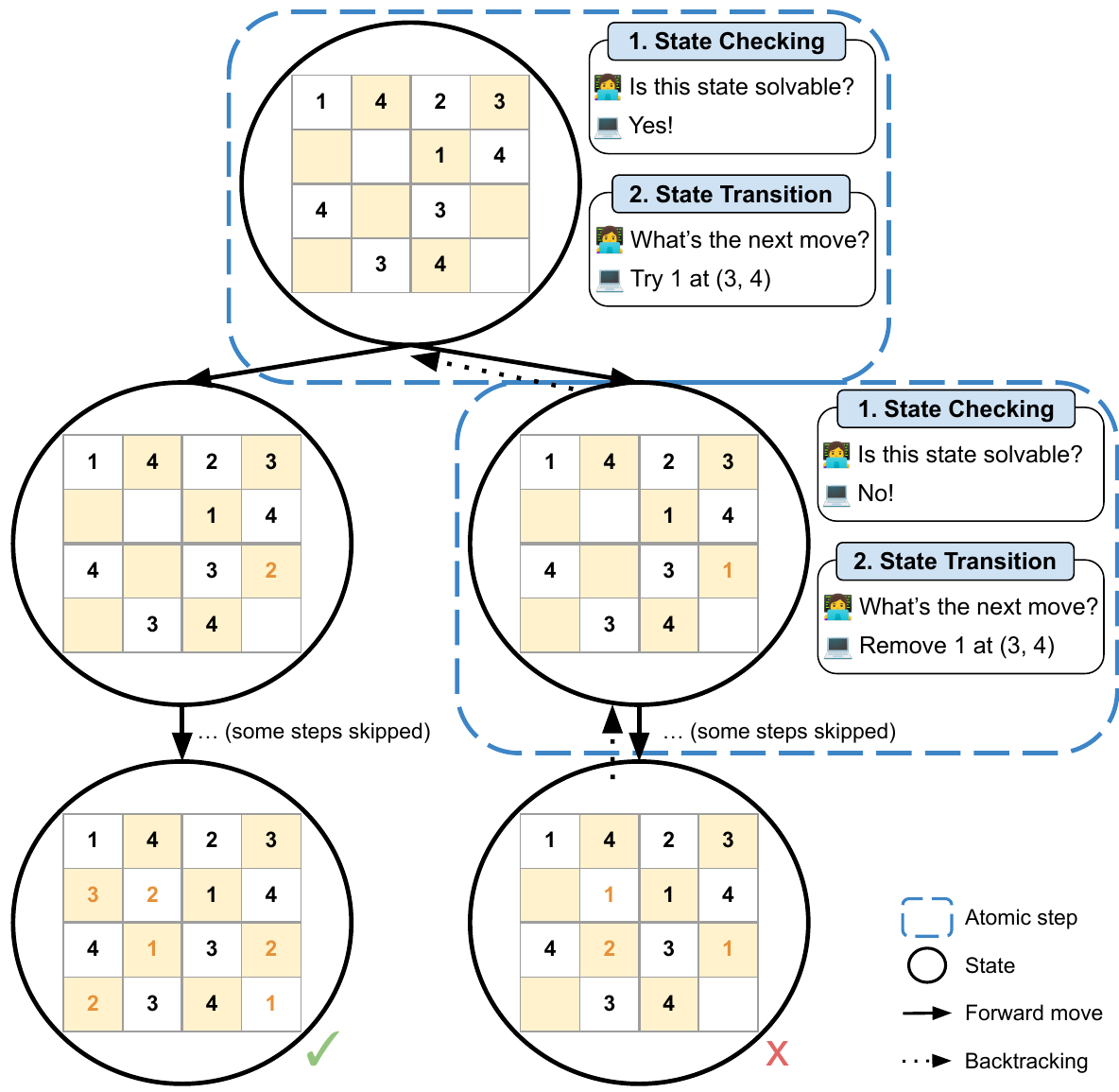}
    \caption{A simplified Sudoku solution tree illustrating stepwise state checking (solvability prediction) and transition (next move determination, including forward moves and backtracking).}
    \label{fig:tree}
\end{figure}

\begin{figure*}[t]
    \centering
    \includegraphics[width=\textwidth]{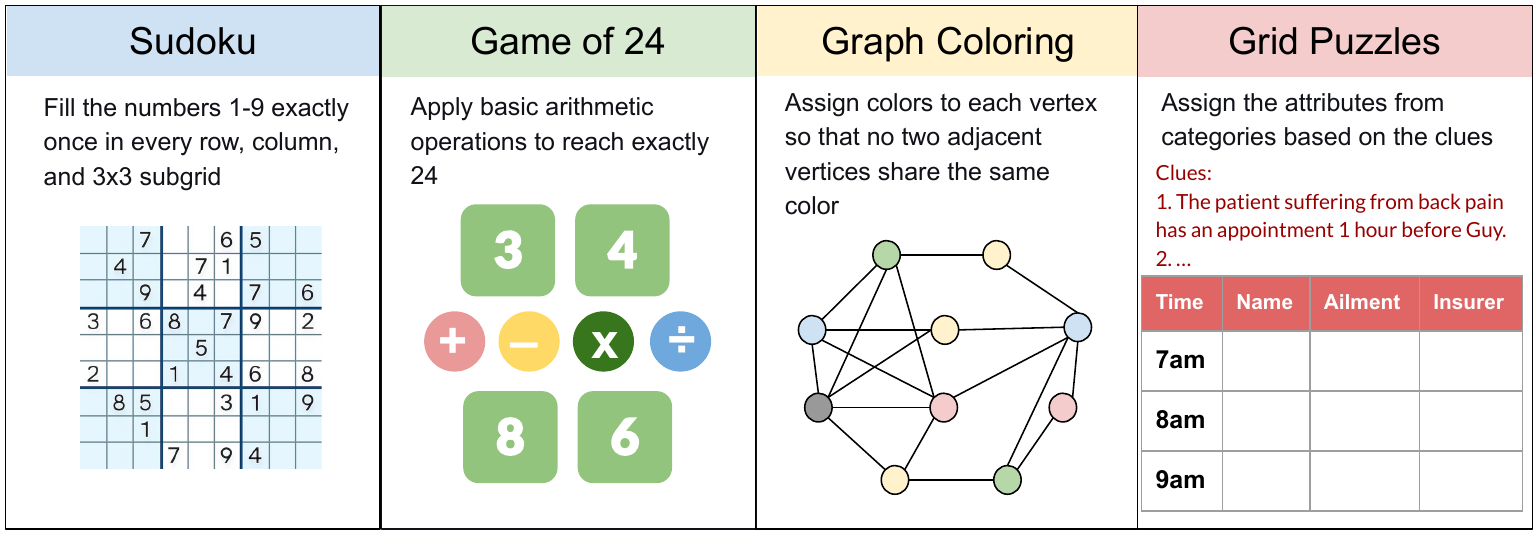}
    \caption{An illustration of four puzzle categories in \framework. These puzzles are solved through discrete steps, with explicit rules allowing easy validation of intermediate states.
    }
    \label{fig:dataset}
\end{figure*}

In cognitive science, human reasoning is typically characterized by two distinct systems: (i) System 1, which is fast, automatic, and effortless, and (ii) System 2, which is slow, analytical, and effortful~\citep{daniel2011@thinking-fast-slow}. System 2 reasoning enables humans to proactively anticipate future outcomes, reassess intermediate states, and iteratively refine solutions~\cite{yao2023tree}, thereby allowing them to tackle more complex reasoning tasks. Recent studies suggest that large language models (LLMs) can attain advantages akin to those of System 2 reasoning~\cite{o1,snell2024scaling,deepseekai2025deepseekr1incentivizingreasoningcapability,kimik15}. Instead of generating answers directly, these models can iteratively reflect and correct their reasoning~\cite{shinn2023reflexion}, achieving strong performance on reasoning-intensive tasks like mathematics and coding~\cite{qin2024o1, li2024can, muennighoff2025s1}.

Despite these improvements, the underlying mechanisms remain underexplored. Existing reasoning benchmarks primarily focus on the final-answer accuracy~\cite{hendrycks2021measuring, cobbe2021training, chen2021evaluating, math}, which offers limited insight into LLMs’ internal reasoning processes, such as reflection and correction. For instance, models might reach a correct conclusion through flawed reasoning~\cite{zelikman2022star,creswell2023selectioninference,lightman2024lets,chia-etal-2024-reasoning}. This diminishes the trustworthiness of model outputs, posing potential risks in practical usage. Moreover, models might achieve high accuracy by exploiting superficial patterns in the training data~\cite{roelofs2019meta,xu-etal-2023-peerda,DBLP:journals/corr/SpuriousCorrelations24}, making it difficult to ascertain whether the observed performance gain truly stems from deliberate reasoning. Therefore, there is a pressing need for more comprehensive reasoning benchmarks that assess the integrity of intermediate processes.

In this work, we present \framework, a logic-puzzle benchmark for evaluating LLMs' deliberate reasoning capabilities, such as reflection, backtracking, and exploring alternative solutions. \framework includes four types of puzzles: \textit{Sudoku}, \textit{Graph Coloring}, \textit{Game of 24}, and \textit{Grid Puzzles}, as shown in Figure \ref{fig:dataset}. Solving a logic puzzle involves a series of discrete steps, and the explicit rules make it straightforward to validate the intermediate states. To structure our evaluation, we propose two key actions in each atomic step, \textbf{state checking} and \textbf{state transition}, as illustrated in Figure~\ref{fig:tree}. State checking predicts whether the current state can lead to a solvable solution~\cite{agarwal2019model,wang2024lookahead}. It captures both retrospective evaluation of prior steps~\cite{lightman2024lets} and prospective analysis of future steps. Meanwhile, state transition focuses on determining the next valid step, either moving forward or backtracking to the previous state. Together, these two tasks cover the entire puzzle-solving process, revealing the internal reasoning processes of reflection, correction, and exploration of alternative paths in LLMs.

Our evaluation reveals a significant 19.7\% performance gap between OpenAI-o1~\cite{o1} and Gemini-2.0-Flash-Thinking~\cite{geminiflash}, a difference not captured by other maths and coding benchmarks where high scores saturate. General-purpose models like GPT-4o~\cite{gpt4o} struggle with deliberate reasoning in \framework, making near-random guesses in state checking and poor performance in state transition.

To enhance reasoning, we develop a specialized training set on puzzle state checking and transition. Integrating our dataset with common reasoning data consistently improves performance on complex reasoning tasks. For example, when applied to DeepSeek-R1-Distill-Qwen-7B, our puzzle data boosts the accuracy from 82.3\% to 87.4\% on GSM8K, compared to models trained exclusively on math data. Our results suggest that skills such as backtracking and constraint validation generalize from puzzles to general reasoning, similar to how structured practice (e.g., chess) enhances human strategic thinking and problem solving.

Our main contributions are three-fold: 1) We introduce \framework, a puzzle-based benchmark accompanied by systematic evaluations on state checking and transition, to provide a more precise evaluation of models' reasoning abilities, particularly in reflection and correction. 2) Experimental results reveal substantial limitations in deliberate reasoning among general-purpose models, and even in the leading reasoning models. 3) We show that training on structured puzzles transfers deliberate reasoning skills to general problem-solving.

\section{\framework}

\begin{table*}[ht]
  \centering
  \resizebox{\linewidth}{!}{
  \begin{tabular}{llll}
    \hline
    \textbf{Puzzle}           & \textbf{Puzzle State}  &  \textbf{Minimal Move} \\
    \hline
    Sudoku & Partial / Complete 9x9 board  &  Add / Remove a digit          \\
    Graph Coloring & A graph of partially / completely colored vertex & Color / Uncolor a vertex                      \\
    Game of 24 & Partial / Complete arithmetic expression  & Apply / Unapply an operation to two remaining numbers  \\
    Logic Grid Puzzles & Partial / Complete grid & Assign / Remove attributes according to a given clue \\
    \hline
  \end{tabular}
  }
  \caption{The definition of minimal move for each category of logic puzzles in our \framework.}
  \label{tab:node-edge-meaning} 
\end{table*}

We present \framework, a logic-puzzle benchmark that comprehensively assesses LLMs' reasoning through stepwise evaluation of state checking and transition. Inspired by the adage ``think twice before acting,'' these actions capture how models assess the current situation (\ie state checking) and plan the next move (\ie state transition), skills crucial for deliberate reasoning.

Formally, the reasoning process of a logic puzzle can be represented as $p=\{p_1, p_2,...,p_n\}$, where $n$ denotes the number of atomic steps. Each step $p_i$ consists of a puzzle state $s_i$ and two actions: state checking $a^c_i$ and state transition $a^t_i$,  \ie $p_i=(s_i, a^c_i, a^t_i)$. Applying these actions to $s_i$ produces the next state $s_{i+1}$. The puzzle-solving process begins at an initial state $s_1$ and proceeds through a sequence of atomic steps until reaching the solution state $s_n$ that satisfies all constraints. 

In the following section, we first introduce a tree-based approach for step decomposition in our puzzles (\S\ref{sec:construction}). Next, we describe the two key actions -- state checking and state transition -- that facilitate reasoning evaluation of models (\S\ref{sec:evaluation_methods}).

\subsection{Tree-based Puzzle Decomposition}
\label{sec:construction}

Puzzle solving can be represented as a tree, where nodes are intermediate states, and edges represent the execution of state checking and state transition, as illustrated in Figure~\ref{fig:tree}. Edges are bidirectional, enabling both the exploration of child states and backtracking to the parent when encountering dead ends. This process begins at the root node $s_1$ and terminates at a solution leaf $s_n$, potentially requiring multiple backtracks to explore different paths.

To capture all possible states, we perform a depth-first search (DFS) from the initial puzzle state $s_1$ until no further valid states remain for exploration. Each DFS step involves a minimal move to ensure that no valid state is overlooked. For example, in Sudoku, we add or remove only a single digit at each step. Table~\ref{tab:node-edge-meaning} summarizes the definition of a minimal move for each puzzle category. Additionally, we translate puzzle rules into executable code to automatically validate each state.

\paragraph{Sudoku.} 
In Sudoku, the aim is to fill the empty cells such that each row, each column, and each of the $3\times3$ subgrids contains all digits from 1 to 9 exactly once. A Sudoku state can be either a partially or fully filled $9\times 9$ grid. The minimal move is defined as either adding a number for exploration or removing a number for backtracking. We use Sudoku questions from Kaggle\footnote{\url{https://www.kaggle.com/datasets/bryanpark/sudoku}} to create the Sudoku trees.

\paragraph{Graph Coloring.} 
The aim of graph coloring is to assign colors to each vertex in a graph such that no two adjacent vertices share the same color. Each puzzle specifies a maximum number of colors allowed in a graph. A graph coloring state is either a partially colored graph or a completely colored graph. A minimal move involves either assigning a color to a vertex or removing a color from a vertex. To create the questions, we generate random graphs and find their respective chromatic numbers using the backtracking algorithm~\cite{backtracking}.

\paragraph{Game of 24.}
In Game of 24, given four numbers, the task is to apply basic arithmetic operations (addition, subtraction, multiplication, and division) to reach exactly the value of 24. Each number must be used exactly once. Each state is a partial or complete arithmetic expression. The minimal move is to apply or unapply an operation to two of the remaining numbers. We use the problem set from \citet{yao2023tree} to generate the trees.

\paragraph{Logic Grid Puzzles.}
Logic grid puzzles involve filling a grid with attributes from multiple categories (\eg color, time) based on a set of textual clues. Each attribute must appear exactly once per category and satisfy all given clues. Each state is a partially or fully filled grid, with minimal moves being adding or removing attribute combinations specified in a clue. Our grid puzzle trees are constructed from \citet{tyagi-etal-2024-step}. Unlike other puzzles, translating textual clues into verification code is challenging, especially when it involves attribute comparisons. To address this, we define three functions: $r(v)$ and $c(v)$ to retrieve the row and column numbers of an attribute $v$, and $T(i,j)$ to identify the attribute at row $i$ and column $j$. These functions encode attributes to a structured axis space. Thus, the textual clues can be parsed into conditions involving  $r(v)$, $c(v)$, and $T(i,j)$ for constraint checks. For example, Clue 1 in Figure~\ref{fig:dataset} can be parsed to $T(r($``Guy''$), c($``Time''$)) - T(r($``back pain''$), c($``Time''$)) == 1$. We use one-shot prompting with GPT-4o~\cite{gpt4o} for clue translation, followed by manual verification. We ensure all solutions pass the coded clues.

\subsection{Evaluation Tasks}
\label{sec:evaluation_methods}

 We define two key actions, state checking and state transition, which enable movements between states.

\paragraph{State Checking.}
State checking requires the model to assess if a given state $s_i$ can lead to a solvable solution $s_n$. Based on our constructed puzzle trees, we label a state as solvable if at least one valid solution exists in the subtree of $s_i$. To ensure a diverse difficulty range, we uniformly sample both solvable and unsolvable states across different tree depths. For each sampled state, we prompt models to evaluate its solvability with puzzle rules, prior visited states, and the current state (see Appendix~\ref{sec:prompts}). In general, state checking evaluates two key aspects: 1) It tests if models can reflect on prior steps to avoid rule violations (e.g., preventing duplicate ``1''s in a Sudoku row). 2) It requires models to anticipate potential dead ends by looking ahead. The second aspect, however, requires a higher level of reasoning to proactively detect unsolvable states.

\paragraph{State Transition.}
State transition involves making the minimal move defined in \S\ref{sec:construction}, which requires models to determine the next valid state. Based on the state-checking results, models should proceed if the state is solvable and backtrack otherwise. Specifically, at a solvable state, a correct transition would be an unvisited child of the given state. At an unsolvable state, the correct move is to backtrack to its parent state. To isolate the impact of state transition from incorrect state checking, our evaluation provides ground-truth state-checking labels. We sample states from the puzzle tree and construct prompts with puzzle rules, prior visited states, the current state, and some unsolvable child states (see Appendix~\ref{sec:prompts}). The inclusion of unsolvable child states is to assess whether models can effectively bypass these bad states.

\begin{table}[t]
\small
\centering
\setlength{\tabcolsep}{3mm}{
\begin{tabular}{l l r}
\toprule
\textbf{Puzzle} & \textbf{Model} & \textbf{End-to-end Acc.}\\
\midrule

\multirow{6}{*}{Sudoku} 

& \cellcolor{lightgray}GPT-4o                   &  \cellcolor{lightgray}0\\
& \cellcolor{lightgray}GPT-3.5            &  \cellcolor{lightgray}0 \\
& \cellcolor{lightgray}Gemini-F & \cellcolor{lightgray}\textbf{5.9} \\
& \cellcolor{lightgray}Qwen2.5-Inst    & \cellcolor{lightgray}0 \\
& \cellcolor{lightgray2}Gemini-FT        & \cellcolor{lightgray2} 0 \\
& \cellcolor{lightgray2}o1                       & \cellcolor{lightgray2} 0\\
\cmidrule(lr){1-3}

\multirow{6}{*}{Graph Coloring}
& \cellcolor{lightgray}GPT-4o                   & \cellcolor{lightgray}7.8 \\
& \cellcolor{lightgray}GPT-3.5            &  \cellcolor{lightgray}3.9 \\
& \cellcolor{lightgray}Gemini-F & \cellcolor{lightgray}35.3 \\
& \cellcolor{lightgray}Qwen2.5-Inst    & \cellcolor{lightgray}2.0 \\
& \cellcolor{lightgray2}Gemini-FT   & \cellcolor{lightgray2} \textbf{80.4}\\
& \cellcolor{lightgray2}o1                       & \cellcolor{lightgray2} 78.4 \\
\cmidrule(lr){1-3}

\multirow{6}{*}{Game of 24}
& \cellcolor{lightgray}GPT-4o                   &  \cellcolor{lightgray}15.3\\
& \cellcolor{lightgray}GPT-3.5            & \cellcolor{lightgray}3.1 \\
& \cellcolor{lightgray}Gemini-F & \cellcolor{lightgray}\textbf{83.7} \\
& \cellcolor{lightgray}Qwen2.5-Inst    &  \cellcolor{lightgray}17.3 \\
& \cellcolor{lightgray2}Gemini-FT         & \cellcolor{lightgray2}48.0  \\
& \cellcolor{lightgray2}o1                       & \cellcolor{lightgray2}54.1 \\
\cmidrule(lr){1-3}

\multirow{6}{*}{Grid Puzzles}
& \cellcolor{lightgray}GPT-4o                   &  \cellcolor{lightgray}2.2 \\
& \cellcolor{lightgray}GPT-3.5            &  \cellcolor{lightgray}2.2 \\
& \cellcolor{lightgray}Gemini-F & \cellcolor{lightgray}10.9 \\
& \cellcolor{lightgray}Qwen2.5-Inst    &  \cellcolor{lightgray}4.4 \\
& \cellcolor{lightgray2}Gemini-FT         & \cellcolor{lightgray2} 34.8 \\
& \cellcolor{lightgray2}o1                       & \cellcolor{lightgray2}\textbf{45.7}  \\

\bottomrule
\end{tabular}
}
\caption{A preliminary study on end-to-end puzzle-solving performance of LLMs.}
\label{tab:e2e-results}
\end{table}

\section{Experimental Setup}

\paragraph{Datasets.}
We sample 500 intermediate states per puzzle category, resulting in 2000 test instances for each task: state checking and state transition. Dataset statistics are included in Appendix~\ref{sec:dataset-statis}.

\paragraph{Implementation.}
We use zero-shot chain of thought prompting~\cite{kojima2022large} for evaluation. To ensure a genuine evaluation of LLM's inherent reasoning capabilities, we explicitly include ``\texttt{Do not solve using programming}'' in the instruction, restricting models from relying on memorized code snippets from their training data. In our actual attempts, without explicitly stating this constraint, models tend to generate Sudoku solvers or graph-coloring algorithms instead of demonstrating deliberate reasoning. Prompt templates are shown in Appendix \ref{sec:prompts}.

\paragraph{Models.}
We select the best-performing open and closed-source models, including 1) reasoning-oriented models: o1~\cite{o1}, Gemini-2.0-Flash-Thinking (Gemini-FT, \citet{geminiflash}), and 2) general-purpose models that perform straightforward execution: GPT-4o~\cite{gpt4o}, GPT-3.5~\cite{gpt3.5}, Gemini-2.0-Flash (Gemini-F, \citet{geminiflash}), and Qwen2.5-72B-Instruct (Qwen2.5-Inst, \citet{qwen25}).

\begin{table}[t]
\small
\centering
\resizebox{\linewidth}{!}{
\begin{tabular}{l l r r r}
\toprule
\textbf{Puzzle} & \textbf{Model} & \textbf{SC Acc.} & \textbf{ST Acc.} & \textbf{Avg.} \\
\midrule

\multirow{7}{*}{Sudoku} 
& Random          & 50.0 & - & - \\
& \cellcolor{lightgray}GPT-4o                   
& \cellcolor{lightgray}52.4 
& \cellcolor{lightgray}38.8 
& \cellcolor{lightgray}45.6 \\
& \cellcolor{lightgray}GPT-3.5            
& \cellcolor{lightgray}49.0   
& \cellcolor{lightgray}10.6 
& \cellcolor{lightgray}29.8 \\
& \cellcolor{lightgray}Gemini-F         
& \cellcolor{lightgray}50.4 
& \cellcolor{lightgray}39.0 
& \cellcolor{lightgray}44.7 \\
& \cellcolor{lightgray}Qwen2.5-Inst    
& \cellcolor{lightgray}51.6 
& \cellcolor{lightgray}26.6 
& \cellcolor{lightgray}39.1 \\
& \cellcolor{lightgray2}Gemini-FT 
& \cellcolor{lightgray2}69.2 
& \cellcolor{lightgray2}48.8 
& \cellcolor{lightgray2}59.0\\
& \cellcolor{lightgray2}o1                       
& \cellcolor{lightgray2}\textbf{81.0}   
& \cellcolor{lightgray2}\textbf{70.2} 
& \cellcolor{lightgray2}\textbf{75.6} \\
\cmidrule(lr){1-5}

\multirow{7}{*}{Graph Coloring}
& Random          & 50.0 & - & - \\
& \cellcolor{lightgray}GPT-4o                   & \cellcolor{lightgray}56.4 & 
\cellcolor{lightgray}49.4 & 
\cellcolor{lightgray}52.9 \\
& 
\cellcolor{lightgray}GPT-3.5            & 
\cellcolor{lightgray}52.2 & 
\cellcolor{lightgray}20.4 & 
\cellcolor{lightgray}36.3 \\
& \cellcolor{lightgray}Gemini-F         & 
\cellcolor{lightgray}56.8 & 
\cellcolor{lightgray}45.8 & 
\cellcolor{lightgray}51.3 \\
& \cellcolor{lightgray}Qwen2.5-Inst    & 
\cellcolor{lightgray}58.6 & 
\cellcolor{lightgray}35.4 & 
\cellcolor{lightgray}47.0 \\
& \cellcolor{lightgray2}Gemini-FT & 
\cellcolor{lightgray2}92.6 & 
\cellcolor{lightgray2}46.4 & 
\cellcolor{lightgray2}69.5 \\
& \cellcolor{lightgray2}o1                       & \cellcolor{lightgray2}\textbf{94.6} & 
\cellcolor{lightgray2}\textbf{65.0}   & 
\cellcolor{lightgray2}\textbf{79.8} \\
\cmidrule(lr){1-5}

\multirow{7}{*}{Game of 24}
&Random          &50.0 & - & - \\

& \cellcolor{lightgray}GPT-4o                   & \cellcolor{lightgray}82.6 & \cellcolor{lightgray}23.0 & \cellcolor{lightgray}52.8 \\
& \cellcolor{lightgray}GPT-3.5            & \cellcolor{lightgray}56.4 & \cellcolor{lightgray}14.2 & \cellcolor{lightgray}35.3 \\
& \cellcolor{lightgray}Gemini-F         & \cellcolor{lightgray}93.4 & \cellcolor{lightgray}54.6 & \cellcolor{lightgray}74.0 \\
& \cellcolor{lightgray}Qwen2.5-Inst    & \cellcolor{lightgray}88.2 & \cellcolor{lightgray}39.2 & \cellcolor{lightgray}63.7 \\
& \cellcolor{lightgray2}Gemini-FT & \cellcolor{lightgray2}96.0 & \cellcolor{lightgray2}48.6 & \cellcolor{lightgray2}72.3 \\
& \cellcolor{lightgray2}o1                       & \cellcolor{lightgray2}\textbf{97.4} & \cellcolor{lightgray2}\textbf{86.6} & \cellcolor{lightgray2}\textbf{92.0} \\
\cmidrule(lr){1-5}

\multirow{7}{*}{Grid Puzzles}
& Random          & 50.0 & - & - \\
& \cellcolor{lightgray}GPT-4o                   & \cellcolor{lightgray}52.4 & \cellcolor{lightgray}10.0 & \cellcolor{lightgray}31.2 \\
& \cellcolor{lightgray}GPT-3.5            & \cellcolor{lightgray}42.6 & \cellcolor{lightgray}11.4 & \cellcolor{lightgray}27.0 \\
& \cellcolor{lightgray}Gemini-F         & \cellcolor{lightgray}37.4 & \cellcolor{lightgray}18.8 & \cellcolor{lightgray}28.1 \\
& \cellcolor{lightgray}Qwen2.5-Inst    & \cellcolor{lightgray}40.8 & \cellcolor{lightgray}9.6  & \cellcolor{lightgray}25.2 \\
& \cellcolor{lightgray2}Gemini-FT & \cellcolor{lightgray2}\textbf{89.0} & \cellcolor{lightgray2}51.4 & \cellcolor{lightgray2}70.2 \\
& \cellcolor{lightgray2}o1                       & \cellcolor{lightgray2}88.8 & \cellcolor{lightgray2}\textbf{77.6} & \cellcolor{lightgray2}\textbf{83.2} \\
\bottomrule
\end{tabular}
}
\caption{The state checking and transition accuracy using \framework, where \textbf{SC} and \textbf{ST} denote state checking and transition, respectively.}
\label{tab:evaluation-results}
\end{table}

\section{Evaluation Results}\label{sec:eval-results}

In this section, we first present a preliminary evaluation of LLMs on end-to-end puzzle solving (\S\ref{sec:preliminary}), which reveals inconsistencies in model performance. To gain deeper insights beyond their end-to-end performance, we present our main results focusing on state checking and transition tasks (\S\ref{sec:main-results}), followed by analyses on models' behaviors, error patterns, and performance across different difficulty levels (\S\ref{sec:sc-breakdown} to \S\ref{sec:performance-by-difficulty}).

\subsection{Preliminary: End-to-End Puzzle Solving}\label{sec:preliminary}
Table~\ref{tab:e2e-results} presents an initial evaluation of LLMs on end-to-end puzzle-solving tasks. Despite their strong performance on coding and math tasks~\cite{qwen25}, these models are notably weak in end-to-end puzzle solving. Additionally, there are some counter-intuitive observations: Gemini-F outperforms Gemini-FT on Sudoku and Game of 24, yet struggles on the other two puzzles. These inconsistencies suggest that end-to-end puzzle solving alone may not be a reliable metric for assessing LLMs’ reasoning, emphasizing the need for more granular evaluation methods.

\subsection{Main results}\label{sec:main-results}
To understand models' reasoning capabilities in greater depth, we evaluate on state checking and transition. Results in Table~\ref{tab:evaluation-results} reveal noticeable performance gaps between reasoning-oriented and general-purpose models. On the state-checking task, reasoning-oriented models (o1 and Gemini-FT) consistently lead the performance in every puzzle category. In contrast, general-purpose models barely match the random baseline in puzzles like Sudoku and Grid Puzzles. A similar trend is observed on the state-transition task. These findings further support the view that inference-time scaling can substantially boost reasoning capabilities~\cite{snell2024scaling,muennighoff2025s1}.

Between the reasoning models, Gemini-FT generally performs on par with o1 in state checking but consistently lags behind in state transition. This reveals weaknesses in Gemini-FT's reasoning process, particularly in error revision. These findings align well with our practical experience using these models, which provides empirical evidence that \framework offers a more accurate reflection of LLMs' reasoning capabilities. 

\subsection{Model Behaviors in State Checking}
\label{sec:sc-breakdown}

State checking requires looking ahead to identify unsolvable states. To analyze models' behaviors, we report the state-checking precision, recall, and F1 scores in Table~\ref{tab:f-scores}, designating unsolvable states as positive cases. Recall measures the model's ability to detect dead ends, while precision indicates the reliability of its unsolvable state predictions.

We find that reasoning models generally detect unsolvable states well, as indicated by the high F1 scores. As for general models (GPT-4o, Gemini-F, Qwen2.5-Inst), the recall is generally low in Sudoku but not in Game of 24, possibly due to Sudoku's deeper puzzle tree making unsolvable state detection harder. A similar trend is also observed in other puzzles that have deeper solutions, including graph coloring and grid puzzles (see Appendix~\ref{sec:f-scores-full}). These observations reveal that general LLMs tend to make overly optimistic decisions (i.e., assuming a solvable state) when faced with tasks that exceed their actual capabilities.

Nevertheless, GPT-4o and Qwen2.5-Inst show high precision, which suggests that these models are conservative and might not attempt to classify states as unsolvable unless they are very confident.

\begin{table}[!t]
\small
\centering
\resizebox{1.0\linewidth}{!}{
\begin{tabular}{llrrrr}
\toprule
\textbf{Puzzle} & \textbf{Model} & \textbf{Recall} & \textbf{Precision} & \textbf{F1} \\
\midrule
\multirow{6}{*}{Sudoku}
& GPT-4o & 6.4 & 80.0   & 11.9 \\
& GPT-3.5 & 28.0  & 49.0   & 35.6 \\
& Gemini-F & 3.2 & 57.1 & 6.06 \\
& Qwen2.5-Inst & 4.8 & 75.0   & 9.02 \\
& Gemini-FT & \textbf{87.2} & 64.3 & 74.0 \\
& o1 & 73.2 & \textbf{86.7} & \textbf{79.4} \\
\midrule
\multirow{6}{*}{Game 24} 
& GPT-4o & 95.6 & 75.9 & 84.6 \\
& GPT-3.5 & 54.8	& 56.6 &	55.7 \\
& Gemini-F & 98.8 &	89.2 &	93.7 \\
& Qwen2.5-Inst & 97.6 &	82.2 &	89.2 \\
& Gemini-FT & 94.8 & 97.5 & 96.1 \\
& o1 & \textbf{95.6} & \textbf{99.2} & \textbf{97.4} \\

\bottomrule
\end{tabular}
}
\caption{\label{tab:f-scores}Precision, recall, and F1 scores of state-checking task in \framework.}
\end{table}

\subsection{Quality Analysis of State Checking}
\label{sec:sc-error}

To examine the errors made in state checking, we conduct a human analysis of the mistakes from the best-performing model, o1, in Figure~\ref{fig:error_types}. The most common error is the Misinterpretation of Premises, where o1 incorrectly uses available information to reach a faulty conclusion. For instance, in a grid puzzle, given the clue ``The chocolate piece, Joey's cake, and the \$125 cake are three different cakes,'' the model still mistakenly assigned Joey's cake to the \$125 cake. Additionally, the model might fail to explore alternative paths, leading to an incorrect assessment of current states. Other mistakes include drawing a wrong conclusion despite correct deductions (Inconsistent Reasoning), failure to recognize conflicting information (Conflicts Resolving Failure), nonexistent constraints (False Premise), and a few instruction-following errors. Examples of the errors are shown in Appendix \ref{sec:sc-errors}.

\begin{figure}
\centering
\resizebox{1.0\linewidth}{!}{
\begin{tikzpicture}
\pie[rotate=0, text = legend, color={blue!30, red!30, orange!30, violet!27, yellow!30, gray}]
 {39.2/Misinterpretation of Premises, 
 17.6/Exploration and Backtracking Issues, 
 15.7/Inconsistent Reasoning,
 13.7/Conflicts Resolving Failure,
 11.8/False Premise,
 2/Others
} 
   
    \end{tikzpicture}
    }
\caption{Human analysis of error types.}
\label{fig:error_types}
\end{figure}

\pgfplotstableread{
Label Parent Sibling Multiple Invalid Child
o1 86	59	1	1	0
Gemini-FT 109 65 1 20 0
GPT-4o 157	7 7	107	1
GPT-3.5 146	63		76	50	97
Gemini-F 139	37		1	124	4
Qwen 215	21		0	121	1
    }\testdata
\begin{figure*}[t!]
    \centering
    \begin{subfigure}[b]{0.5\textwidth}
        \centering
        \begin{tikzpicture}
        \begin{axis}[
            width=\textwidth,
            height=5.8cm,
            ybar=0cm,
            bar width=9pt,
            ymax=100,
            ymin=0,
            xtick = {1,2,3,4,5,6},
            xticklabels = {o1, Gemini-FT, GPT-4o, GPT-3.5,  Gemini-F, Qwen},
            ylabel = {Accuracy (\%)},
            xtick pos = left,
            ytick pos = left,
            tickwidth=0mm,
            ymajorgrids = true,
            ytick={0,20,40,60,80,100},
            enlarge x limits=0.12,
            grid style=dashed, 
            legend style={at={(1,1)}, anchor=north east, legend columns=-1, font=\scriptsize},  
            xticklabel style={font=\scriptsize},
            yticklabel style={font=\scriptsize},
            ylabel style={font=\scriptsize}
        ]
        \addplot coordinates {(1,98.8) (2,77.2) (3,49.6) (4,5.2)  (5,48.4) (6,48.8)};    
        \addlegendentry{Solvable};
        \addplot coordinates {(1,41.6) (2,20.4) (3,28) (4,16) (5,29.6) (6,4.4)};    
        \addlegendentry{Unsolvable};

    \legend{Solvable, Unsolvable}
    \end{axis}
    \end{tikzpicture}
    \vspace{-2mm}
    \caption{State-transition performance breakdown by class}
    \label{fig:ST-breakdown-by-class}
    \end{subfigure}%
    \hfill
    \begin{subfigure}[b]{0.5\textwidth}
        \centering
        \begin{tikzpicture}

    \begin{axis}[
        ybar stacked,
        width=\textwidth,
        height=5.8cm,
        ymin=0,
        ymax=600,
        bar width=12pt,
        xtick=data,
        ytick={0,250,500},
        ylabel = {Count},
        ylabel style={font=\scriptsize},
        legend style={at={(1,1)}, anchor=north east, legend columns=3, font=\scriptsize}, 
        reverse legend=true, 
        xticklabels from table={\testdata}{Label},
        xticklabel style={font=\scriptsize},
        yticklabel style={font=\scriptsize}]
    \addplot  table [y=Sibling, meta=Label, x expr=\coordindex] {\testdata};
    \addlegendentry{Sibling}
    \addplot  table [y=Parent, meta=Label, x expr=\coordindex] {\testdata};
    \addlegendentry{Backtracking failure}
    
    \addplot  table [y=Child, meta=Label, x expr=\coordindex] {\testdata};
    \addlegendentry{Unsolvable child}
    \addplot  table [y=Invalid, meta=Label, x expr=\coordindex] {\testdata};
    \addlegendentry{Invalid move}

    \addplot  table [y=Multiple, meta=Label, x expr=\coordindex] {\testdata};
    \addlegendentry{Multiple moves}

    \end{axis}
    \end{tikzpicture}
    \vspace{-2mm}
    \caption{State-transition error type breakdown}
    \label{fig:ST-error}
    \end{subfigure}
    
    \caption{Performance breakdown and error analysis of state-transition in Sudoku.
    }
    \label{fig:ST-breakdown}
\end{figure*}

\subsection{Model Behaviors in State Transition}
\label{sec:st-breakdown}

To understand models' behaviors during state transition, we break down the performance by class and count the common mistakes made by models.

The left chart of Figure~\ref{fig:ST-breakdown} shows the Sudoku state-transition performance breakdown for each class (solvable vs. unsolvable). We observe that most models transit much better on solvable states than on unsolvable ones. The large gap indicates that models are better at proceeding forward from a valid state than backtracking. This might be attributed to a forward‐generation reasoning style of LLMs. This trend, however, does not apply to GPT-3.5, which shows significantly weak performance and tends to rely primarily on random guessing.

The right chart of Figure~\ref{fig:ST-breakdown} shows the errors typically made by models during state transition. At solvable states, common errors include making multiple moves (Multiple Moves), violating Sudoku rules (Invalid Move), and transitioning to an unsolvable child state (Unsolvable Child). At unsolvable states, two primary errors are: failing to return to the parent state (Backtracking Failure) and making an additional move to a sibling state after backtracking (Sibling). Examples of the errors are shown in Appendix \ref{sec:st-errors}. Among these errors, Backtracking Failure is the most frequent across all models. Models sometimes jump back more than one level (\eg to a grandparent state) or to a wrong state, indicating that LLMs struggle with step-by-step backtracking. For reasoning models (o1 and Gemini-FT), transitioning to siblings is the second most frequent error. This error is due to violating the minimal move principle (Table~\ref{tab:node-edge-meaning}), highlighting weaknesses in their instruction-following capability. For general models, they frequently commit an invalid move. This shows that general models often fail to adequately check Sudoku rules.

\subsection{Performance by Difficulty Level}
\label{sec:performance-by-difficulty}

To understand the state-checking performance across difficulty levels, we plot the density diagrams of correct vs. incorrect predictions by the number of unfilled cells in Sudoku states (Figure \ref{fig:difficulty}). We find that Sudoku states with fewer empty cells are more likely to be predicted correctly. As the number of unfilled cells increases, the problem becomes more complex and requires more exploration, and the proportion of incorrect predictions increases. This likely reflects the increased computational complexity of looking ahead and determining solvability when many possibilities exist.

\begin{figure}[t!]
 \centering
    \includegraphics[width=0.95\linewidth]{assets/density-plot.pdf}
    \caption{Density plot of number of empty cells for correct vs. incorrect predictions.
    }
    \label{fig:difficulty}
\end{figure}

\section{Training on puzzle data}
As highlighted in \S\ref{sec:eval-results}, most models struggle with identifying dead ends and backtracking. These observations reveal critical bottlenecks in models' reflection and correction abilities, which are essential for reasoning. We hypothesize that training on state-checking and state-transition tasks can provide synthetic reflection, allow models to practice error correction, and ultimately transfer these reasoning skills to mathematical problem solving.

To validate our hypothesis, we construct a training set consisting of state-checking and state-transition data from Sudoku, Graph Coloring, and Game of 24. We train models on a mixture of puzzle data and math data to test whether reasoning skills transfer beyond puzzles themselves, ultimately improving performance on math tasks.

\subsection{Experimental setup}
We prepare $10,000$ puzzle samples and another $15,000$ samples from MetaMathQA~\citep{yu2024metamath}, a popular training set for mathematical reasoning. The puzzle states are easily verified, making it suitable for Reinforcement Learning. Specifically, we train GRPO~\citep{shao2024deepseekmathpushinglimitsmathematical} on DeepSeek-R1-Distill-Qwen-1.5B and DeepSeek-R1-Distill-Qwen-7B~\citep{deepseekai2025deepseekr1incentivizingreasoningcapability}. Other training details are reported in Appendix~\ref{sec:training-details}. Due to limited computing resources, we restrict the maximum completion length to 1024 in both training and evaluation.

\subsection{Improvements on mathematical reasoning}
We start with $2,000$ training samples, with a $1:1$ distribution between puzzle and math data. We compare with two baselines: the first is the base model, and the second is training with entirely math samples. The results in Table~\ref{tab:main-training} show that combining puzzle data with math data yields the highest accuracy on both GSM8K and MATH for both models, outperforming training on math data alone. This consistent performance improvement suggests that the state-checking and state-transition logic of puzzle solving generalizes to mathematical problems, aligning well with our initial hypothesis. 

\begin{table}[!t]
\small
\centering
\resizebox{1.0\linewidth}{!}{
\begin{tabular}{llrr}
\toprule
\textbf{Model} & \textbf{Data} & \textbf{GSM8K} & \textbf{MATH} \\
\midrule
\multirow{3}{*}{DeepSeek-R1-Distill-Qwen-1.5B} & None & 65.5 & 45.6 \\
& Math-only & 73.6 & 51.1 \\
& Mixed & \textbf{76.1} & \textbf{53.1} \\

\midrule
\multirow{3}{*}{DeepSeek-R1-Distill-Qwen-7B} & None & 79.7 & 63.2 \\
& Math-only & 82.3 & 70.7 \\
& Mixed & \textbf{87.4} & \textbf{71.4} \\

\bottomrule
\end{tabular}
}
\caption{\label{tab:main-training} Training with our puzzle data improves math reasoning on GSM8K and MATH.}
\end{table}

To further assess the impact of the mixed training, we analyze the number of correctly solved samples at each difficulty level in the MATH dataset, where a higher level means higher difficulty. Results in Appendix \ref{sec:math-difficulty-analysis} indicate that mixed training is especially effective for harder problems (levels 4 and 5), likely because complex problems require more reflection and correction. By incorporating puzzle-solving data into math reasoning training, we encourage models to reflect and backtrack when necessary.

To investigate the nature of improvements, we examine corrected examples after training. Around 60\% of these examples use longer reasoning steps, determined by sentence splitter. Through quality analysis of 20 examples, we find more evidence of verification and self-correction in the thought process. Examples are shown in Appendix \ref{sec:mixed-training-quality}.

\begin{figure}
\centering
\begin{tikzpicture}
\pgfplotsset{width = \linewidth, height = 5cm}
    \begin{axis}[
        ymax=80,
        ymin=70,
        ylabel={GSM8K Accuracy (\%)},
        xlabel={Proportion of math samples ($r_{m}$)},
        xtick = {0,4,5,6,7,8,9,10},
        xticklabels = {0,0.4,0.5,0.6,0.7,0.8,0.9,1},
        xtick pos = left,
        ytick pos = left,
        ymajorgrids = true,
        grid style=dashed,
        xticklabel style={font=\scriptsize},
        xlabel style={font=\scriptsize},
        yticklabel style={font=\scriptsize},
        ylabel style={font=\scriptsize}
    ]
    \addplot [mark=*,mark options={solid}, dashed, color=gray, thick]
    plot coordinates {
    (0, 72.6) (4, 74.6) };
    \addplot [mark=*, color=gray, thick]
    plot coordinates {
    (4, 74.6) (5, 75.4) (6, 76.7) (7, 77.3) (8, 79) (9, 76) (10, 77) };
    \end{axis}
\end{tikzpicture}
\caption{The effect of math ratio on the 1.5B model.}
\label{fig:ratio}
\end{figure}

\begin{table}[!t]
\small
\centering
\resizebox{0.6\linewidth}{!}{
\begin{tabular}{lr}
\toprule
\textbf{Data} & \textbf{GSM8K} \\
\midrule
None & 65.5 \\
Math-only & 73.6 \\
Mixed ($r_{u}=0.2$) & 74.8\\
Mixed ($r_{u}=0.5$) & 76.1\\
Mixed ($r_{u}=0.8$) & 77.4\\
\bottomrule
\end{tabular}
}
\caption{The effect of unsolvable ratio on the 1.5B model.}
\label{tab:unsolvable-ratio} 
\end{table}

\subsection{Effect of the puzzle ratio}
To study the optimal ratio between puzzle-based and math-based data, we vary the proportion of math samples, $r_{m}$,  from $0.4$ to $1.0$ in a combined training set of $10k$ samples. In Figure~\ref{fig:ratio}, the performance on GSM8K steadily improves as the math ratio increases, peaking at a ratio of  $0.8$. Beyond this point, increasing the math ratio further results in lower accuracy. This suggests that neither pure math training nor pure puzzle training is optimal. Instead, a balanced combination of puzzle-based and math data provides the best performance. This indicates that our puzzle-based data, though simple, can complement the reasoning in standard math problems.

In addition, our previous analysis shows that models struggle significantly more with unsolvable states than solvable states in both state checking (\S\ref{sec:sc-breakdown}) and state transition (\S\ref{sec:st-breakdown}). Our error analysis also shows that “backtracking failure” is a major error (\S\ref{sec:sc-error}), even for top reasoning models like o1. This means we could prioritize the learning for backtracking from these unsolvable cases to maximize learning benefits. To validate this, we examine the effect of varying the ratio of unsolvable data, $r_{u}$, in mixed training. In Table \ref{tab:unsolvable-ratio}, it is evident that increasing the proportion of unsolvable data boosts the performance on GSM8K, suggesting that focusing on these unsolvable cases enhances the model’s ability to detect dead ends and revise strategies—skills transferable to math problem-solving.

\subsection{Analysis on the scaling effect}

We examine the scaling effect of using math-only and mixed data for training. We use the optimal math ratio ($r_m = 0.8$) in the mixed data. The results in Figure \ref{fig:scaling} show that, as we increase the number of training samples, both approaches benefit from scaling up. Noticeably, the mixed approach consistently outperforms math-only training starting from $5k$ samples. While math-only training shows diminishing returns or even a slight decline beyond $7.5k$ samples, the mixed approach continues to improve,  reaching an accuracy peak of $81.3\%$ with around $12.5k$ samples. This scaling effect suggests the great potential of our simple puzzle data for enhancing the overall reasoning capability of LLMs.

\begin{figure}
\centering
\begin{tikzpicture}
\pgfplotsset{width = \linewidth, height = 5cm}
    \begin{axis}[
        ymax=83,
        ymin=73,
        ylabel={GSM8K Accuracy (\%)},
        xlabel={Number of training samples (k)},
        xtick = {1,2,3,4,5,6},
        xticklabels = {2.5,5,7.5,10,12.5,15},
        xtick pos = left,
        ytick pos = left,
        ymajorgrids = true,
        grid style=dashed,
        xticklabel style={font=\scriptsize},
        xlabel style={font=\scriptsize},
        yticklabel style={font=\scriptsize},
        ylabel style={font=\scriptsize},
        legend style={at={(0.5,1)}, anchor=north, legend columns=2, font=\scriptsize}, 
    ]
    \addplot [mark=*, dashed, color=red!55, thick]
    plot coordinates {
    (1, 76.5) (2, 77.4) (3, 77.5) (4, 77) (5, 77.3) (6, 75.3) };
    \addlegendentry{Math-only}
    \addplot [mark=*, color=blue!55, thick]
    plot coordinates {
    (1, 75.1) (2, 77.9) (3, 79.3) (4, 79) (5, 81.3) (6, 80.4)};
    \addlegendentry{Mixed}
    \end{axis}
\end{tikzpicture}
\caption{The scaling performance of the 1.5B model.}
\label{fig:scaling}
\end{figure}

\subsection{Improvements on general reasoning}

To investigate if our puzzle data benefits general reasoning tasks beyond math, we conduct additional experiments on MMLU~\citep{hendrycks2021measuring}, a dataset covering 57 tasks including STEM, humanities, social sciences, and others. Using the same optimal ratio ($r_m = 0.8$) as in math training, we observe consistent performance improvements over MMLU-only training (Table \ref{tab:mmlu}).

Next, we analyze the performance improvement across different subsets in MMLU. Results in Appendix \ref{sec:mmlu-analysis} show that mixed training significantly improves performance on complex, multi-step reasoning tasks that might require more reflection (STEM +10\%, Social Sciences +3.8\%), but shows less benefit for simpler, more direct question types.

Overall, our puzzle data bridges the gap of the lack of annotated intermediate training data for general reasoning tasks by providing synthetic, structured scenarios for learning reasoning skills. While real-world reasoning is less deterministic, our results indicate that mastering deterministic steps in puzzle solving builds foundational skills like reflection and correction, which demonstrably transfer to general reasoning tasks.

\begin{table}[!t]
\small
\centering
\resizebox{1.0\linewidth}{!}{
\begin{tabular}{llrr}
\toprule
\textbf{Model} & \textbf{Data} & \textbf{MMLU-test} \\
\midrule
\multirow{3}{*}{DeepSeek-R1-Distill-Qwen-1.5B} & None & 37.8 \\
& MMLU-train-only & 41.1 \\
& Mixed & \textbf{44.9} \\

\midrule
\multirow{3}{*}{DeepSeek-R1-Distill-Qwen-7B} & None & 54.9 \\
& MMLU-train-only & 61.2 \\
& Mixed & \textbf{64.2} \\

\bottomrule
\end{tabular}
}
\caption{\label{tab:mmlu} Training with our puzzle data improves general reasoning tasks on MMLU.}
\end{table}

\section{Related Work}
\paragraph{Reasoning Capabilities of LLMs.}
Advancing the reasoning capabilities of large language models is a critical goal in natural language processing~\cite{wos1992automated,yang2018hotpotqa}. In recent years, LLMs, combined with prompting techniques such as Chain of Thought~\cite{wei2022chain}, Tree of Thought~\cite{yao2023tree}, and Self-Consistency~\cite{wang2023selfconsistency}, have shown remarkable performance in various reasoning tasks~\cite{cobbe2021training,srivastava2022beyond}. Current evaluation methods focus mainly on the final accuracy in reasoning-intensive domains, including mathematics~\cite{cobbe2021training,math,chen-etal-2023-theoremqa,rein2023gpqa,VisAidMath}, science~\cite{mihaylov-etal-2018-suit,xu-etal-2021-exploiting-reasoning,xu-etal-2021-dynamic,DBLP:journals/tkde/HuangCFQZJCJ25,DBLP:conf/acl/HuangZC0WZCJ24}, coding~\cite{chen2021evaluating,austin2021program}, commonsense~\cite{hendrycks2021measuring}, and logical reasoning~\cite{yao2023tree,long2023large}. However, as inference-time scaling gains importance~\cite{snell2024scaling,deepseekai2025deepseekr1incentivizingreasoningcapability} and models become more capable of reasoning, it is crucial to assess how effectively models perform reflection and correction during reasoning. While \citet{tyagi-etal-2024-step} manually analyzes the reasoning chains in logic puzzles, their approach lacks scalability. Some studies~\cite{singh2024exposing,zeng2024mrben,xu-etal-2024-reasons} evaluate how models handle reasoning mistakes, but these investigations often rely on rule-based mistakes that may be easily resolved by current LLMs. Moreover, these studies only assess reflection on past steps in a static manner. In our work, we address these limitations by introducing two novel tasks designed to more accurately reflect models' capabilities in dynamic reasoning and error correction.

\paragraph{Puzzle Solving Tasks.}
Logic puzzles, which require deducing solutions from a set of rules~\cite{giadikiaroglou-etal-2024-puzzle}, are ideal for evaluating the reasoning abilities of LLMs, as they rely minimally on prior knowledge~\cite{li-etal-2024-assessing-logical}. Recent studies have explored LLMs on various puzzles with different emphases~\cite{mittal2024puzzlebench}, such as Sudoku~\cite{ishay2023leveraging,long2023large} for strategic thinking, Game of 24~\cite{ding2023everything,yao2023tree} for arithmetic calculations. Some investigate grid puzzles~\cite{dziri2024faith,tyagi-etal-2024-step}, crosswords~\cite{yao2023tree}, chess puzzles~\cite{feng2024chessgpt}, mazes~\cite{Noever2021PuzzleSW}, Minesweeper~\cite{li-etal-2024-assessing-logical}, and abstract pattern recognition~\cite{chia-etal-2024-puzzlevqa}.
However, the evaluation remains mainly focused on the final accuracy.

\section{Conclusion}
In this work, we introduce \framework, a novel logic-puzzle benchmark designed to comprehensively evaluate the reasoning capabilities of LLMs. Unlike existing benchmarks that focus mainly on final-answer accuracy, \framework delves into intermediate reasoning steps, specifically emphasizing state checking and transition actions. This fine-grained evaluation captures the model's ability to reflect, look ahead, and backtrack, which are vital aspects of human-like System 2 reasoning. Our experiments reveal significant gaps between reasoning-oriented and general-purpose LLMs, emphasizing the need to consider reflection and correction for robust reasoning evaluation. Furthermore, using puzzle-based data for training can enhance performance in general reasoning tasks, highlighting the scalability of this approach and its potential to transfer reasoning skills beyond puzzles to broader reasoning.

\section*{Acknowledgment}
This research is supported by DAMO Academy through DAMO Academy Research Intern Program and Alibaba-NTU Singapore Joint Research Institute (JRI), Nanyang Technological University, Singapore. We would also like to extend our gratitude to Interdisciplinary Graduate Programme and College of Computing and Data Science of NTU, for their support.

\section*{Limitations}
Our study has several limitations. First, we represent puzzle states using textual tables. Our evaluation shows that models can reasonably understand this table format. However, there is potential to explore alternative representation formats, such as coordinates or images. The image format could be particularly valuable for facilitating the evaluation of multi-modal reasoning, offering a promising direction for future extensions of our work. Second, we employ zero-shot CoT prompting~\cite{kojima2022large} to focus on evaluating the inherent reasoning capabilities of LLMs, foregoing more advanced prompting techniques that could potentially improve performance but might obscure the models' true reasoning abilities. Finally, our current evaluation prioritizes reasoning correctness over efficiency, treating all solvable states equally regardless of the steps required to reach the solution. Future research could incorporate metrics like "steps to solution" to assess efficiency.

\bibliography{custom}

\begin{thebibliography}{59}
\providecommand{\natexlab}[1]{#1}

\bibitem[{Agarwal et~al.(2019)Agarwal, Muelling, and Fragkiadaki}]{agarwal2019model}
Arpit Agarwal, Katharina Muelling, and Katerina Fragkiadaki. 2019.
\newblock \href {https://www.researchgate.net/publication/335685785_Model_Learning_for_Look-Ahead_Exploration_in_Continuous_Control} {Model learning for look-ahead exploration in continuous control}.
\newblock In \emph{Proceedings of the AAAI Conference on Artificial Intelligence}, volume~33, pages 3151--3158.

\bibitem[{Austin et~al.(2021)Austin, Odena, Nye, Bosma, Michalewski, Dohan, Jiang, Cai, Terry, Le et~al.}]{austin2021program}
Jacob Austin, Augustus Odena, Maxwell Nye, Maarten Bosma, Henryk Michalewski, David Dohan, Ellen Jiang, Carrie Cai, Michael Terry, Quoc Le, et~al. 2021.
\newblock \href {https://www.researchgate.net/publication/353970323_Program_Synthesis_with_Large_Language_Models} {Program synthesis with large language models}.
\newblock \emph{arXiv preprint arXiv:2108.07732}.

\bibitem[{Chen et~al.(2021)Chen, Tworek, Jun, Yuan, Pinto, Kaplan, Edwards, Burda, Joseph, Brockman et~al.}]{chen2021evaluating}
Mark Chen, Jerry Tworek, Heewoo Jun, Qiming Yuan, Henrique Ponde De~Oliveira Pinto, Jared Kaplan, Harri Edwards, Yuri Burda, Nicholas Joseph, Greg Brockman, et~al. 2021.
\newblock \href {https://www.researchgate.net/publication/353066484_Evaluating_Large_Language_Models_Trained_on_Code} {Evaluating large language models trained on code}.
\newblock \emph{arXiv preprint arXiv:2107.03374}.

\bibitem[{Chen et~al.(2023)Chen, Yin, Ku, Lu, Wan, Ma, Xu, Wang, and Xia}]{chen-etal-2023-theoremqa}
Wenhu Chen, Ming Yin, Max Ku, Pan Lu, Yixin Wan, Xueguang Ma, Jianyu Xu, Xinyi Wang, and Tony Xia. 2023.
\newblock \href {https://doi.org/10.18653/v1/2023.emnlp-main.489} {{T}heorem{QA}: A theorem-driven question answering dataset}.
\newblock In \emph{Proceedings of the 2023 Conference on Empirical Methods in Natural Language Processing}, pages 7889--7901, Singapore. Association for Computational Linguistics.

\bibitem[{Chia et~al.(2024{\natexlab{a}})Chia, Chen, Xu, Luu, Poria, and Bing}]{chia-etal-2024-reasoning}
Yew~Ken Chia, Guizhen Chen, Weiwen Xu, Anh~Tuan Luu, Soujanya Poria, and Lidong Bing. 2024{\natexlab{a}}.
\newblock \href {https://doi.org/10.18653/v1/2024.findings-emnlp.977} {Reasoning paths optimization: Learning to reason and explore from diverse paths}.
\newblock In \emph{Findings of the Association for Computational Linguistics: EMNLP 2024}, pages 16763--16780, Miami, Florida, USA. Association for Computational Linguistics.

\bibitem[{Chia et~al.(2024{\natexlab{b}})Chia, Toh, Ghosal, Bing, and Poria}]{chia-etal-2024-puzzlevqa}
Yew~Ken Chia, Vernon Toh, Deepanway Ghosal, Lidong Bing, and Soujanya Poria. 2024{\natexlab{b}}.
\newblock \href {https://doi.org/10.18653/v1/2024.findings-acl.962} {{P}uzzle{VQA}: Diagnosing multimodal reasoning challenges of language models with abstract visual patterns}.
\newblock In \emph{Findings of the Association for Computational Linguistics: ACL 2024}, pages 16259--16273, Bangkok, Thailand. Association for Computational Linguistics.

\bibitem[{Cobbe et~al.(2021)Cobbe, Kosaraju, Bavarian, Chen, Jun, Kaiser, Plappert, Tworek, Hilton, Nakano et~al.}]{cobbe2021training}
Karl Cobbe, Vineet Kosaraju, Mohammad Bavarian, Mark Chen, Heewoo Jun, Lukasz Kaiser, Matthias Plappert, Jerry Tworek, Jacob Hilton, Reiichiro Nakano, et~al. 2021.
\newblock \href {https://arxiv.org/abs/2110.14168} {Training verifiers to solve math word problems}.
\newblock \emph{arXiv preprint arXiv:2110.14168}.

\bibitem[{Creswell et~al.(2023)Creswell, Shanahan, and Higgins}]{creswell2023selectioninference}
Antonia Creswell, Murray Shanahan, and Irina Higgins. 2023.
\newblock \href {https://openreview.net/forum?id=3Pf3Wg6o-A4} {Selection-inference: Exploiting large language models for interpretable logical reasoning}.
\newblock In \emph{The Eleventh International Conference on Learning Representations}.

\bibitem[{DeepSeek-AI et~al.(2025)DeepSeek-AI, Guo, Yang, Zhang et~al.}]{deepseekai2025deepseekr1incentivizingreasoningcapability}
DeepSeek-AI, Daya Guo, Dejian Yang, Haowei Zhang, et~al. 2025.
\newblock \href {https://arxiv.org/abs/2501.12948} {Deepseek-r1: Incentivizing reasoning capability in llms via reinforcement learning}.
\newblock \emph{Preprint}, arXiv:2501.12948.

\bibitem[{Ding et~al.(2024)Ding, Zhang, Wang, Xu, Ma, Zhang, Qin, Rajmohan, Lin, and Zhang}]{ding2023everything}
Ruomeng Ding, Chaoyun Zhang, Lu~Wang, Yong Xu, Minghua Ma, Wei Zhang, Si~Qin, Saravan Rajmohan, Qingwei Lin, and Dongmei Zhang. 2024.
\newblock \href {https://doi.org/10.18653/v1/2024.findings-acl.95} {Everything of thoughts: Defying the law of penrose triangle for thought generation}.
\newblock In \emph{Findings of the Association for Computational Linguistics: ACL 2024}, pages 1638--1662, Bangkok, Thailand. Association for Computational Linguistics.

\bibitem[{Dziri et~al.(2023)Dziri, Lu, Sclar, Li, Jiang, Lin, Welleck, West, Bhagavatula, Bras, Hwang, Sanyal, Ren, Ettinger, Harchaoui, and Choi}]{dziri2024faith}
Nouha Dziri, Ximing Lu, Melanie Sclar, Xiang~Lorraine Li, Liwei Jiang, Bill~Yuchen Lin, Sean Welleck, Peter West, Chandra Bhagavatula, Ronan~Le Bras, Jena~D. Hwang, Soumya Sanyal, Xiang Ren, Allyson Ettinger, Zaid Harchaoui, and Yejin Choi. 2023.
\newblock \href {https://openreview.net/forum?id=Fkckkr3ya8} {Faith and fate: Limits of transformers on compositionality}.
\newblock In \emph{Thirty-seventh Conference on Neural Information Processing Systems}.

\bibitem[{Enstr{\"{o}}m et~al.(2024)Enstr{\"{o}}m, Kjellberg, and Johansson}]{DBLP:journals/corr/SpuriousCorrelations24}
Daniel Enstr{\"{o}}m, Viktor Kjellberg, and Moa Johansson. 2024.
\newblock \href {https://doi.org/10.48550/ARXIV.2403.11314} {Reasoning in transformers - mitigating spurious correlations and reasoning shortcuts}.
\newblock \emph{CoRR}, abs/2403.11314.

\bibitem[{Feng et~al.(2023)Feng, Luo, Wang, Tang, Yang, Shao, Mguni, Du, and Wang}]{feng2024chessgpt}
Xidong Feng, Yicheng Luo, Ziyan Wang, Hongrui Tang, Mengyue Yang, Kun Shao, David~Henry Mguni, Yali Du, and Jun Wang. 2023.
\newblock \href {https://openreview.net/forum?id=pvdm4B6JMK} {Chess{GPT}: Bridging policy learning and language modeling}.
\newblock In \emph{Thirty-seventh Conference on Neural Information Processing Systems Datasets and Benchmarks Track}.

\bibitem[{Giadikiaroglou et~al.(2024)Giadikiaroglou, Lymperaiou, Filandrianos, and Stamou}]{giadikiaroglou-etal-2024-puzzle}
Panagiotis Giadikiaroglou, Maria Lymperaiou, Giorgos Filandrianos, and Giorgos Stamou. 2024.
\newblock \href {https://doi.org/10.18653/v1/2024.emnlp-main.646} {Puzzle solving using reasoning of large language models: A survey}.
\newblock In \emph{Proceedings of the 2024 Conference on Empirical Methods in Natural Language Processing}, pages 11574--11591, Miami, Florida, USA. Association for Computational Linguistics.

\bibitem[{Google(2024)}]{geminiflash}
Google. 2024.
\newblock \href {https://deepmind.google/technologies/gemini/flash/} {Gemini-2.0-flash family}.

\bibitem[{Hendrycks et~al.(2021{\natexlab{a}})Hendrycks, Burns, Basart, Zou, Mazeika, Song, and Steinhardt}]{hendrycks2021measuring}
Dan Hendrycks, Collin Burns, Steven Basart, Andy Zou, Mantas Mazeika, Dawn Song, and Jacob Steinhardt. 2021{\natexlab{a}}.
\newblock \href {https://openreview.net/forum?id=d7KBjmI3GmQ} {Measuring massive multitask language understanding}.
\newblock In \emph{International Conference on Learning Representations}.

\bibitem[{Hendrycks et~al.(2021{\natexlab{b}})Hendrycks, Burns, Kadavath, Arora, Basart, Tang, Song, and Steinhardt}]{math}
Dan Hendrycks, Collin Burns, Saurav Kadavath, Akul Arora, Steven Basart, Eric Tang, Dawn Song, and Jacob Steinhardt. 2021{\natexlab{b}}.
\newblock \href {https://openreview.net/forum?id=7Bywt2mQsCe} {Measuring mathematical problem solving with the {MATH} dataset}.
\newblock In \emph{Thirty-fifth Conference on Neural Information Processing Systems Datasets and Benchmarks Track (Round 2)}.

\bibitem[{Huang et~al.(2025)Huang, Chan, Fung, Qiu, Zhou, Joty, Chang, and Ji}]{DBLP:journals/tkde/HuangCFQZJCJ25}
Kung{-}Hsiang Huang, Hou~Pong Chan, May Fung, Haoyi Qiu, Mingyang Zhou, Shafiq Joty, Shih{-}Fu Chang, and Heng Ji. 2025.
\newblock \href {https://doi.org/10.1109/TKDE.2024.3513320} {From pixels to insights: {A} survey on automatic chart understanding in the era of large foundation models}.
\newblock \emph{{IEEE} Trans. Knowl. Data Eng.}, 37(5):2550--2568.

\bibitem[{Huang et~al.(2024)Huang, Zhou, Chan, Fung, Wang, Zhang, Chang, and Ji}]{DBLP:conf/acl/HuangZC0WZCJ24}
Kung{-}Hsiang Huang, Mingyang Zhou, Hou~Pong Chan, Yi~Fung, Zhenhailong Wang, Lingyu Zhang, Shih{-}Fu Chang, and Heng Ji. 2024.
\newblock \href {https://doi.org/10.18653/V1/2024.FINDINGS-ACL.41} {Do lvlms understand charts? analyzing and correcting factual errors in chart captioning}.
\newblock In \emph{Findings of the Association for Computational Linguistics, {ACL} 2024, Bangkok, Thailand and virtual meeting, August 11-16, 2024}, pages 730--749. Association for Computational Linguistics.

\bibitem[{Ishay et~al.(2023)Ishay, Yang, and Lee}]{ishay2023leveraging}
Adam Ishay, Zhun Yang, and Joohyung Lee. 2023.
\newblock \href {https://arxiv.org/abs/2307.07699} {Leveraging large language models to generate answer set programs}.
\newblock \emph{arXiv preprint arXiv:2307.07699}.

\bibitem[{Kahneman(2011)}]{daniel2011@thinking-fast-slow}
Daniel Kahneman. 2011.
\newblock \emph{Thinking, fast and slow}.
\newblock Farrar, Straus and Giroux.

\bibitem[{Kojima et~al.(2022)Kojima, Gu, Reid, Matsuo, and Iwasawa}]{kojima2022large}
Takeshi Kojima, Shixiang~Shane Gu, Machel Reid, Yutaka Matsuo, and Yusuke Iwasawa. 2022.
\newblock \href {https://openreview.net/forum?id=e2TBb5y0yFf} {Large language models are zero-shot reasoners}.
\newblock In \emph{Advances in Neural Information Processing Systems}.

\bibitem[{Li et~al.(2024{\natexlab{a}})Li, Xu, Zhao, Jiao, Joty, and Bing}]{li2024can}
Xingxuan Li, Weiwen Xu, Ruochen Zhao, Fangkai Jiao, Shafiq Joty, and Lidong Bing. 2024{\natexlab{a}}.
\newblock \href {https://arxiv.org/abs/2410.01428} {Can we further elicit reasoning in llms? critic-guided planning with retrieval-augmentation for solving challenging tasks}.
\newblock \emph{arXiv preprint arXiv:2410.01428}.

\bibitem[{Li et~al.(2024{\natexlab{b}})Li, Wang, and Zhang}]{li-etal-2024-assessing-logical}
Yinghao Li, Haorui Wang, and Chao Zhang. 2024{\natexlab{b}}.
\newblock \href {https://doi.org/10.18653/v1/2024.naacl-long.4} {Assessing logical puzzle solving in large language models: Insights from a minesweeper case study}.
\newblock In \emph{Proceedings of the 2024 Conference of the North American Chapter of the Association for Computational Linguistics: Human Language Technologies (Volume 1: Long Papers)}, pages 59--81, Mexico City, Mexico. Association for Computational Linguistics.

\bibitem[{Lightman et~al.(2024)Lightman, Kosaraju, Burda, Edwards, Baker, Lee, Leike, Schulman, Sutskever, and Cobbe}]{lightman2024lets}
Hunter Lightman, Vineet Kosaraju, Yuri Burda, Harrison Edwards, Bowen Baker, Teddy Lee, Jan Leike, John Schulman, Ilya Sutskever, and Karl Cobbe. 2024.
\newblock \href {https://openreview.net/forum?id=v8L0pN6EOi} {Let's verify step by step}.
\newblock In \emph{The Twelfth International Conference on Learning Representations}.

\bibitem[{Long(2023)}]{long2023large}
Jieyi Long. 2023.
\newblock \href {https://openreview.net/forum?id=a648X9AoL4} {Large language model guided tree-of-thought}.

\bibitem[{Ma et~al.(2024)Ma, Zhan, Wong, Li, Sun, Chan, and Chao}]{VisAidMath}
Jingkun Ma, Runzhe Zhan, Derek~F. Wong, Yang Li, Di~Sun, Hou~Pong Chan, and Lidia~S. Chao. 2024.
\newblock \href {https://doi.org/10.48550/ARXIV.2410.22995} {Visaidmath: Benchmarking visual-aided mathematical reasoning}.
\newblock \emph{CoRR}, abs/2410.22995.

\bibitem[{Mihaylov et~al.(2018)Mihaylov, Clark, Khot, and Sabharwal}]{mihaylov-etal-2018-suit}
Todor Mihaylov, Peter Clark, Tushar Khot, and Ashish Sabharwal. 2018.
\newblock \href {https://doi.org/10.18653/v1/D18-1260} {Can a suit of armor conduct electricity? a new dataset for open book question answering}.
\newblock In \emph{Proceedings of the 2018 Conference on Empirical Methods in Natural Language Processing}, pages 2381--2391, Brussels, Belgium. Association for Computational Linguistics.

\bibitem[{Mittal et~al.(2024)Mittal, Kartik, Singla et~al.}]{mittal2024puzzlebench}
Chinmay Mittal, Krishna Kartik, Parag Singla, et~al. 2024.
\newblock \href {https://arxiv.org/abs/2402.02611} {Puzzlebench: Can llms solve challenging first-order combinatorial reasoning problems?}
\newblock \emph{arXiv preprint arXiv:2402.02611}.

\bibitem[{Muennighoff et~al.(2025)Muennighoff, Yang, Shi, Li, Fei-Fei, Hajishirzi, Zettlemoyer, Liang, Cand{\`e}s, and Hashimoto}]{muennighoff2025s1}
Niklas Muennighoff, Zitong Yang, Weijia Shi, Xiang~Lisa Li, Li~Fei-Fei, Hannaneh Hajishirzi, Luke Zettlemoyer, Percy Liang, Emmanuel Cand{\`e}s, and Tatsunori Hashimoto. 2025.
\newblock \href {https://arxiv.org/abs/2501.19393} {s1: Simple test-time scaling}.
\newblock \emph{arXiv preprint arXiv:2501.19393}.

\bibitem[{Noever and Burdick(2021)}]{Noever2021PuzzleSW}
David~A. Noever and Ryerson Burdick. 2021.
\newblock \href {https://api.semanticscholar.org/CorpusID:237431487} {Puzzle solving without search or human knowledge: An unnatural language approach}.
\newblock \emph{ArXiv}, abs/2109.02797.

\bibitem[{OpenAI(2022)}]{gpt3.5}
OpenAI. 2022.
\newblock \href {https://platform.openai.com/docs/models/gpt-3-5} {Gpt3.5 turbo}.

\bibitem[{OpenAI(2024)}]{o1}
OpenAI. 2024.
\newblock \href {https://openai.com/index/learning-to-reason-with-llms/} {Learning to reason with llms}.

\bibitem[{OpenAI et~al.(2024)OpenAI, Hurst, Lerer, Goucher et~al.}]{gpt4o}
OpenAI, Aaron Hurst, Adam Lerer, Adam~P. Goucher, et~al. 2024.
\newblock \href {https://arxiv.org/abs/2410.21276} {{GPT}-4o system card}.
\newblock \emph{ArXiv}, abs/2410.21276.

\bibitem[{Qin et~al.(2024)Qin, Li, Zou, Liu, Xia, Huang, Ye, Yuan, Liu, Li et~al.}]{qin2024o1}
Yiwei Qin, Xuefeng Li, Haoyang Zou, Yixiu Liu, Shijie Xia, Zhen Huang, Yixin Ye, Weizhe Yuan, Hector Liu, Yuanzhi Li, et~al. 2024.
\newblock \href {https://arxiv.org/abs/2410.18982} {O1 replication journey: A strategic progress report--part 1}.
\newblock \emph{arXiv preprint arXiv:2410.18982}.

\bibitem[{Qwen et~al.(2025)Qwen, Yang, Zhang, Hui, Zheng, Yu, Li, Liu, Huang, Wei, Lin, Yang, Tu, Zhang, Yang, Yang, Zhou, Lin, Dang, Lu, Bao, Yang, Yu, Li, Xue, Zhang, Zhu, Men, Lin, Li, Tang, Xia, Ren, Ren, Fan, Su, Zhang, Wan, Liu, Cui, Zhang, and Qiu}]{qwen25}
An~Yang Qwen, Baosong Yang, Beichen Zhang, Binyuan Hui, Bo~Zheng, Bowen Yu, Chengyuan Li, Dayiheng Liu, Fei Huang, Haoran Wei, Huan Lin, Jian Yang, Jianhong Tu, Jianwei Zhang, Jianxin Yang, Jiaxi Yang, Jingren Zhou, Junyang Lin, Kai Dang, Keming Lu, Keqin Bao, Kexin Yang, Le~Yu, Mei Li, Mingfeng Xue, Pei Zhang, Qin Zhu, Rui Men, Runji Lin, Tianhao Li, Tianyi Tang, Tingyu Xia, Xingzhang Ren, Xuancheng Ren, Yang Fan, Yang Su, Yichang Zhang, Yu~Wan, Yuqiong Liu, Zeyu Cui, Zhenru Zhang, and Zihan Qiu. 2025.
\newblock \href {https://arxiv.org/abs/2412.15115} {Qwen2.5 technical report}.
\newblock \emph{ArXiv}, abs/2412.15115.

\bibitem[{Rein et~al.(2024)Rein, Hou, Stickland, Petty, Pang, Dirani, Michael, and Bowman}]{rein2023gpqa}
David Rein, Betty~Li Hou, Asa~Cooper Stickland, Jackson Petty, Richard~Yuanzhe Pang, Julien Dirani, Julian Michael, and Samuel~R. Bowman. 2024.
\newblock \href {https://openreview.net/forum?id=Ti67584b98} {{GPQA}: A graduate-level google-proof q\&a benchmark}.
\newblock In \emph{First Conference on Language Modeling}.

\bibitem[{Roelofs et~al.(2019)Roelofs, Shankar, Recht, Fridovich-Keil, Hardt, Miller, and Schmidt}]{roelofs2019meta}
Rebecca Roelofs, Vaishaal Shankar, Benjamin Recht, Sara Fridovich-Keil, Moritz Hardt, John Miller, and Ludwig Schmidt. 2019.
\newblock \href {https://openreview.net/forum?id=HJlr9NBgUr} {A meta-analysis of overfitting in machine learning}.
\newblock \emph{Advances in Neural Information Processing Systems}, 32.

\bibitem[{Shao et~al.(2024)Shao, Wang, Zhu, Xu, Song, Bi, Zhang, Zhang, Li, Wu, and Guo}]{shao2024deepseekmathpushinglimitsmathematical}
Zhihong Shao, Peiyi Wang, Qihao Zhu, Runxin Xu, Junxiao Song, Xiao Bi, Haowei Zhang, Mingchuan Zhang, Y.~K. Li, Y.~Wu, and Daya Guo. 2024.
\newblock \href {https://arxiv.org/abs/2402.03300} {Deepseekmath: Pushing the limits of mathematical reasoning in open language models}.
\newblock \emph{Preprint}, arXiv:2402.03300.

\bibitem[{Shinn et~al.(2023)Shinn, Cassano, Gopinath, Narasimhan, and Yao}]{shinn2023reflexion}
Noah Shinn, Federico Cassano, Ashwin Gopinath, Karthik~R Narasimhan, and Shunyu Yao. 2023.
\newblock \href {https://openreview.net/forum?id=vAElhFcKW6} {Reflexion: language agents with verbal reinforcement learning}.
\newblock In \emph{Thirty-seventh Conference on Neural Information Processing Systems}.

\bibitem[{Singh et~al.(2024)Singh, Nambi, and Vineet}]{singh2024exposing}
Joykirat Singh, Akshay Nambi, and Vibhav Vineet. 2024.
\newblock \href {https://arxiv.org/abs/2406.10834} {Exposing the achilles' heel: Evaluating llms ability to handle mistakes in mathematical reasoning}.
\newblock \emph{arXiv preprint arXiv:2406.10834}.

\bibitem[{Snell et~al.(2025)Snell, Lee, Xu, and Kumar}]{snell2024scaling}
Charlie~Victor Snell, Jaehoon Lee, Kelvin Xu, and Aviral Kumar. 2025.
\newblock \href {https://openreview.net/forum?id=4FWAwZtd2n} {Scaling {LLM} test-time compute optimally can be more effective than scaling parameters for reasoning}.
\newblock In \emph{The Thirteenth International Conference on Learning Representations}.

\bibitem[{Srivastava et~al.(2023)Srivastava, Rastogi, Rao, Shoeb, Abid, Fisch, Brown, Santoro, Gupta, Garriga-Alonso et~al.}]{srivastava2022beyond}
Aarohi Srivastava, Abhinav Rastogi, Abhishek Rao, Abu Awal~Md Shoeb, Abubakar Abid, Adam Fisch, Adam~R Brown, Adam Santoro, Aditya Gupta, Adri{\`a} Garriga-Alonso, et~al. 2023.
\newblock \href {https://openreview.net/forum?id=uyTL5Bvosj} {Beyond the imitation game: Quantifying and extrapolating the capabilities of language models}.
\newblock \emph{Transactions on Machine Learning Research}.
\newblock Featured Certification.

\bibitem[{Team(2025)}]{kimik15}
Kimi Team. 2025.
\newblock \href {https://arxiv.org/abs/2501.12599} {Kimi k1.5: Scaling reinforcement learning with llms}.
\newblock \emph{ArXiv}, abs/2501.12599.

\bibitem[{Tyagi et~al.(2024)Tyagi, Parmar, Kulkarni, Rrv, Patel, Nakamura, Mitra, and Baral}]{tyagi-etal-2024-step}
Nemika Tyagi, Mihir Parmar, Mohith Kulkarni, Aswin Rrv, Nisarg Patel, Mutsumi Nakamura, Arindam Mitra, and Chitta Baral. 2024.
\newblock \href {https://doi.org/10.18653/v1/2024.emnlp-main.1111} {Step-by-step reasoning to solve grid puzzles: Where do {LLM}s falter?}
\newblock In \emph{Proceedings of the 2024 Conference on Empirical Methods in Natural Language Processing}, pages 19898--19915. Association for Computational Linguistics.

\bibitem[{{van Beek}(2006)}]{backtracking}
Peter {van Beek}. 2006.
\newblock \href {https://doi.org/10.1016/S1574-6526(06)80008-8} {Chapter 4 - backtracking search algorithms}.
\newblock In Francesca Rossi, Peter {van Beek}, and Toby Walsh, editors, \emph{Handbook of Constraint Programming}, volume~2 of \emph{Foundations of Artificial Intelligence}, pages 85--134. Elsevier.

\bibitem[{Wang et~al.(2023)Wang, Wei, Schuurmans, Le, Chi, Narang, Chowdhery, and Zhou}]{wang2023selfconsistency}
Xuezhi Wang, Jason Wei, Dale Schuurmans, Quoc~V Le, Ed~H. Chi, Sharan Narang, Aakanksha Chowdhery, and Denny Zhou. 2023.
\newblock \href {https://openreview.net/forum?id=1PL1NIMMrw} {Self-consistency improves chain of thought reasoning in language models}.
\newblock In \emph{The Eleventh International Conference on Learning Representations}.

\bibitem[{Wang et~al.(2024)Wang, Li, Yang, Liu, Hu, Jiang, and Jiang}]{wang2024lookahead}
Zihan Wang, Xiangyang Li, Jiahao Yang, Yeqi Liu, Junjie Hu, Ming Jiang, and Shuqiang Jiang. 2024.
\newblock \href {https://openaccess.thecvf.com/content/CVPR2024/papers/Wang_Lookahead_Exploration_with_Neural_Radiance_Representation_for_Continuous_Vision-Language_Navigation_CVPR_2024_paper.pdf} {Lookahead exploration with neural radiance representation for continuous vision-language navigation}.
\newblock In \emph{Proceedings of the IEEE/CVF Conference on Computer Vision and Pattern Recognition}, pages 13753--13762.

\bibitem[{Wei et~al.(2022)Wei, Wang, Schuurmans, Bosma, brian ichter, Xia, Chi, Le, and Zhou}]{wei2022chain}
Jason Wei, Xuezhi Wang, Dale Schuurmans, Maarten Bosma, brian ichter, Fei Xia, Ed~H. Chi, Quoc~V Le, and Denny Zhou. 2022.
\newblock \href {https://openreview.net/forum?id=_VjQlMeSB_J} {Chain of thought prompting elicits reasoning in large language models}.
\newblock In \emph{Advances in Neural Information Processing Systems}.

\bibitem[{Wos et~al.(1992)Wos, Overbeek, Lusk, and Boyle}]{wos1992automated}
Larry Wos, Ross Overbeek, Ewing Lusk, and Jim Boyle. 1992.
\newblock \emph{Automated reasoning introduction and applications}.
\newblock McGraw-Hill, Inc.

\bibitem[{Xu et~al.(2024)Xu, Cai, Zhang, Lam, and Shi}]{xu-etal-2024-reasons}
Weiwen Xu, Deng Cai, Zhisong Zhang, Wai Lam, and Shuming Shi. 2024.
\newblock \href {https://doi.org/10.18653/v1/2024.findings-acl.730} {Reasons to reject? aligning language models with judgments}.
\newblock In \emph{Findings of the Association for Computational Linguistics: ACL 2024}, pages 12288--12304, Bangkok, Thailand. Association for Computational Linguistics.

\bibitem[{Xu et~al.(2021{\natexlab{a}})Xu, Deng, Zhang, Cai, and Lam}]{xu-etal-2021-exploiting-reasoning}
Weiwen Xu, Yang Deng, Huihui Zhang, Deng Cai, and Wai Lam. 2021{\natexlab{a}}.
\newblock \href {https://doi.org/10.18653/v1/2021.findings-emnlp.99} {Exploiting reasoning chains for multi-hop science question answering}.
\newblock In \emph{Findings of the Association for Computational Linguistics: EMNLP 2021}, pages 1143--1156, Punta Cana, Dominican Republic. Association for Computational Linguistics.

\bibitem[{Xu et~al.(2023)Xu, Li, Deng, Lam, and Bing}]{xu-etal-2023-peerda}
Weiwen Xu, Xin Li, Yang Deng, Wai Lam, and Lidong Bing. 2023.
\newblock \href {https://doi.org/10.18653/v1/2023.acl-long.484} {{P}eer{DA}: Data augmentation via modeling peer relation for span identification tasks}.
\newblock In \emph{Proceedings of the 61st Annual Meeting of the Association for Computational Linguistics (Volume 1: Long Papers)}, pages 8681--8699, Toronto, Canada. Association for Computational Linguistics.

\bibitem[{Xu et~al.(2021{\natexlab{b}})Xu, Zhang, Cai, and Lam}]{xu-etal-2021-dynamic}
Weiwen Xu, Huihui Zhang, Deng Cai, and Wai Lam. 2021{\natexlab{b}}.
\newblock \href {https://doi.org/10.18653/v1/2021.findings-acl.90} {Dynamic semantic graph construction and reasoning for explainable multi-hop science question answering}.
\newblock In \emph{Findings of the Association for Computational Linguistics: ACL-IJCNLP 2021}, pages 1044--1056, Online. Association for Computational Linguistics.

\bibitem[{Yang et~al.(2018)Yang, Qi, Zhang, Bengio, Cohen, Salakhutdinov, and Manning}]{yang2018hotpotqa}
Zhilin Yang, Peng Qi, Saizheng Zhang, Yoshua Bengio, William Cohen, Ruslan Salakhutdinov, and Christopher~D. Manning. 2018.
\newblock \href {https://doi.org/10.18653/v1/D18-1259} {{H}otpot{QA}: A dataset for diverse, explainable multi-hop question answering}.
\newblock In \emph{Proceedings of the 2018 Conference on Empirical Methods in Natural Language Processing}, pages 2369--2380, Brussels, Belgium. Association for Computational Linguistics.

\bibitem[{Yao et~al.(2023)Yao, Yu, Zhao, Shafran, Griffiths, Cao, and Narasimhan}]{yao2023tree}
Shunyu Yao, Dian Yu, Jeffrey Zhao, Izhak Shafran, Thomas~L. Griffiths, Yuan Cao, and Karthik~R Narasimhan. 2023.
\newblock \href {https://openreview.net/forum?id=5Xc1ecxO1h} {Tree of thoughts: Deliberate problem solving with large language models}.
\newblock In \emph{Thirty-seventh Conference on Neural Information Processing Systems}.

\bibitem[{Yu et~al.(2024)Yu, Jiang, Shi, YU, Liu, Zhang, Kwok, Li, Weller, and Liu}]{yu2024metamath}
Longhui Yu, Weisen Jiang, Han Shi, Jincheng YU, Zhengying Liu, Yu~Zhang, James Kwok, Zhenguo Li, Adrian Weller, and Weiyang Liu. 2024.
\newblock \href {https://openreview.net/forum?id=N8N0hgNDRt} {Metamath: Bootstrap your own mathematical questions for large language models}.
\newblock In \emph{The Twelfth International Conference on Learning Representations}.

\bibitem[{Zelikman et~al.(2022)Zelikman, Wu, Mu, and Goodman}]{zelikman2022star}
Eric Zelikman, Yuhuai Wu, Jesse Mu, and Noah Goodman. 2022.
\newblock \href {https://openreview.net/forum?id=_3ELRdg2sgI} {{ST}ar: Bootstrapping reasoning with reasoning}.
\newblock In \emph{Advances in Neural Information Processing Systems}.

\bibitem[{Zeng et~al.(2024)Zeng, Liu, Wan, Li, Chen, Dai, Yao, Xu, Qi, Zhao, Shen, Lu, Tan, Chen, Zhang, Shi, Wang, Guo, and Jia}]{zeng2024mrben}
Zhongshen Zeng, Yinhong Liu, Yingjia Wan, Jingyao Li, Pengguang Chen, Jianbo Dai, Yuxuan Yao, Rongwu Xu, Zehan Qi, Wanru Zhao, Linling Shen, Jianqiao Lu, Haochen Tan, Yukang Chen, Hao Zhang, Zhan Shi, Bailin Wang, Zhijiang Guo, and Jiaya Jia. 2024.
\newblock \href {https://openreview.net/forum?id=GN2qbxZlni} {{MR}-ben: A meta-reasoning benchmark for evaluating system-2 thinking in {LLM}s}.
\newblock In \emph{The Thirty-eighth Annual Conference on Neural Information Processing Systems}.

\end{thebibliography}
\clearpage
\appendix

\section{Appendix}

\subsection{Dataset Statistics}
\label{sec:dataset-statis}
Table~\ref{tab:dataset_statistics} presents the statistics for the four tasks, including the total number of questions, as well as the number of solvable and unsolvable states for each task. For grid puzzles, we can only sample 94 solvable states with unsolvable children, resulting in a somewhat imbalanced dataset. Nonetheless, we have maintained a balance between solvable and unsolvable states for the remaining three puzzles.

\begin{table}[ht!]
\centering
 \resizebox{\linewidth}{!}{
\begin{tabular}{lccc}
\toprule
\textbf{Task} & \textbf{Questions} & \textbf{Solvable States} & \textbf{Unsolvable States} \\
\midrule
Sudoku & 51 & 250 & 250 \\
Graph Coloring & 51 & 250 & 250 \\
Game 24 & 98 & 250 & 250 \\
Grid Puzzles & 50 & 94 & 406 \\
\bottomrule
\end{tabular}}
\caption{Dataset Statistics}
\label{tab:dataset_statistics}
\end{table}

\subsection{Prompt Templates}
\label{sec:prompts}

In Table~\ref{tab:prompts-sudoku}, ~\ref{tab:prompts-graph}, ~\ref{tab:prompts-game24}, ~\ref{tab:prompts-grid-sc}, ~\ref{tab:prompts-grid-st}, we show the state checking and state transition prompts for each puzzle.

\subsection{Additional results on state checking precision, recall and F1 scores}
\label{sec:f-scores-full}

Table~\ref{tab:f-scores-full} reports the state-checking precision, recall, and F1 scores of models across four tasks. It is observed that o1 consistently outperforms all other models in detecting unsolvable states, as evidenced by the high recall and precision acorss tasks. Models like GPT-4o, Qwen2.5-Inst and Gemini-F are generally more precise when they predict unsolvability, but are limited by low recall in tasks like Sudoku and Grid Puzzles. GPT-3.5 generally struggles with both recall and precision, especially in more complex tasks like Sudoku.

\begin{table}[ht!]
\centering
\small
\resizebox{\linewidth}{!}{
\begin{tabular}{llrrrr}
\toprule
\textbf{Puzzle} & \textbf{Model} & \textbf{Recall} & \textbf{Precision} & \textbf{F1} \\
\midrule
\multirow{6}{*}{Sudoku} &
o1 & 73.2 & 86.7 & 79.4 \\
& GPT-4o & 6.4 & 80.0   & 11.9 \\
& GPT-3.5 & 28.0  & 49.0   & 35.6 \\
& Gemini-FT & 87.2 & 64.3 & 74.0 \\
& Gemini-F & 3.2 & 57.1 & 6.06 \\
& Qwen2.5-Inst & 4.8 & 75.0   & 9.02 \\
\midrule
\multirow{6}{*}{Game of 24} &
o1 & 95.6 & 99.2 & 97.4 \\
& GPT-4o & 95.6 & 75.9 & 84.6 \\
& GPT-3.5 & 54.8	& 56.6 &	55.7 \\
& Gemini-FT & 94.8 & 97.5 & 96.1 \\
& Gemini-F & 98.8 &	89.2 &	93.7 \\
& Qwen2.5-Inst & 97.6 &	82.2 &	89.2 \\
\midrule 
\multirow{6}{*}{Graph Coloring} 
& o1 & 93.1 & 95.9 & 94.5 \\ 
& GPT-4o & 44.8 & 57.8 & 50.5 \\ 
& GPT-3.5 & 27.4 & 53.5 & 36.3 \\ 
& Gemini-FT & 96.8 & 89.2 & 92.8 \\
& Gemini-F & 29.0 & 64.3 & 40.0 \\
& Qwen2.5-Inst & 25.8 & 73.6 & 38.2 \\
\midrule 
\multirow{5}{*}{Grid Puzzles} & o1 & 93.8 & 92.5 & 93.2 \\
& GPT-4o & 47.8 & 88.2 & 62.0 \\
& GPT-3.5 & 39.4 & 79.6 & 52.7 \\
& Gemini-FT & 91.6 & 94.7 & 93.1 \\
& Gemini-F & 24.4 & 94.3 & 38.7 \\
\bottomrule
\end{tabular}
}
\caption{\label{tab:f-scores-full}Precision, Recall and F1 scores of state checking task for all puzzles.}
\end{table}

\subsection{Error types in state checking}
\label{sec:sc-errors}
Through human analysis, we categorise five common types of errors o1 made in state checking.
We show one example for each type of error: Misinterpretation of Premises (Table \ref{tab:misintepretation}), Exploration and Backtracking Issues (Table \ref{tab:exploration}), Inconsistent Reasoning (Table \ref{tab:inconsistent-reasoning}), Conflicts Resolving Failure (Table \ref{tab:conflicts}), and False Premise (Table \ref{tab:false-premise}).

\subsection{Error types in state transition}
\label{sec:st-errors}
We automatically classify the state transition mistakes based on the next move. We show one example for each type of error models made in the state transition: Multiple Moves (Table \ref{tab:multiple-moves-example}), Invalid Move (Table \ref{tab:invalid-move-example}), Unsolvable Child (Table \ref{tab:unsolvable-child-example}), Backtracking Failure (Table \ref{tab:backtracking-failure-example}), Sibling (Table \ref{tab:sibling-example}).

\subsection{Training Details}
\label{sec:training-details}
We train our models using GRPO based on OpenR1\footnote{https://github.com/huggingface/open-r1}. We use one GPU to run vLLM for faster generation and the remaining GPUs for training. The hyperparameters and training details are reported in Table~\ref{tab:params}.

\begin{table}[h]
    \centering
    \small
    \begin{tabular}{lccccccc}
    \toprule
    Learning rate & 4e-5 \\
    Warm up ratio & 0.1 \\
    Batch size & 112 \\
    Max prompt length & 1024 \\
    Max completion length & 1024 \\
    Training epochs & 1 \\
    Hardware & 8 GPUs (120 GB) \\
    \bottomrule
    \end{tabular}
    \caption{Hyperparameter and training details.}
    \label{tab:params}
\end{table}

\subsection{Benefits of mixed-training across difficulty level}
\label{sec:math-difficulty-analysis}

Table \ref{tab:math-difficulty-analysis} analyses the impact of mixed training on performance across varying difficulty levels within the MATH dataset. Notably, the gains achieved through mixed training are most significant for levels 4 and 5 problems, which represent the most challenging instances. This suggests that the benefits of mixed training are amplified when tackling complex problems that necessitate more extensive reflection and iterative correction processes.

\begin{table}[!t]
\small
\centering
\resizebox{0.9\linewidth}{!}{
\begin{tabular}{crrr}
\toprule
\textbf{Difficulty Level} & \textbf{Math-only} & \textbf{Mixed} & \textbf{Difference (\%)} \\
\midrule
Level 1 &	372 &	386 &	4\%\\
Level 2 &	622	& 630 &	1\% \\
Level 3 &	675	& 661	& -2\%\\
Level 4	& 540 &	588	& 9\%\\
Level 5 &	348	& 389	& 12\%\\
\bottomrule
\end{tabular}
}
\caption{The \% improvement of correct samples from mixed training using the 1.5B model.}
\label{tab:math-difficulty-analysis} 
\end{table}

\subsection{Quality analysis of model outputs with/without mixed training}
\label{sec:mixed-training-quality}

Through quality analysis of 20 examples, we found more evidence of verification and self-correction in their thought process. Here we show two examples in Table \ref{tab:math-quality-analysis}.

In the first example, model 1 applied the formula directly without cross-checking its consistency with the given example of the third row. After training, model 2 verified the calculation against the provided example before reaching the answer.

In the second example, model 2 initially included an incorrect letter ``C'' but it caught this mistake and removed “C” before proceeding. However, model 1 skipped this step.

These examples suggest that puzzle training helps reinforce skills like constraint validation and error correction, which generalize to math reasoning.

\subsection{Benefits of mixed-training across different reasoning tasks}
\label{sec:mmlu-analysis}

In Table~\ref{tab:MMLU-analysis}, we analyze the performance improvement across different subsets in MMLU. We organize the tasks into four categories following ~\citet{hendrycks2021measuring}. The most significant improvement is in STEM tasks, where accuracy increases by 10\%. The STEM subset consists of multi-step reasoning problems that require reflection and correction in the reasoning process. We also observe some benefits (+3.8\%) to the social sciences subset, as some subjects (eg, Econometrics) require multi-step reasoning. The humanities subset and other tasks, which are typically straightforward questions requiring fewer reasoning steps, show smaller improvements.

\begin{table}[h]
\small
\centering
\resizebox{1.0\linewidth}{!}{
\begin{tabular}{lrrr}
\toprule
\textbf{Category} & \textbf{MMLU-only} & \textbf{Mixed} & \textbf{Improvement (\%)} \\
\midrule
Humanities &	33.0 &	33.8 &	0.8\\
Social Sciences	& 45.4 &	49.2 &	3.8 \\
STEM &	48.7 &	58.7 &	10.0\\
Others &	41.2 &	43.3 &	2.1\\
\bottomrule
\end{tabular}
}
\caption{The improvement in MMLU performance from mixed training using the 1.5B model across different task categories.}
\label{tab:MMLU-analysis} 
\end{table}

\begin{table*}[ht]
    \centering
    \resizebox{\linewidth}{!}{
	\setlength{\tabcolsep}{0mm}{
            \begin{tabular}{p{20cm}}
            \toprule
            \textbf{Sudoku State Checking} \\
            \midrule
                You are given a partially filled 9x9 Sudoku grid represented as a list of lists, where empty cells are represented as 0. \\
                Your task is to determine if this current state can lead to a solvable solution. Specifically, use lookahead techniques to determine if it's possible to fill the remaining cells according to standard Sudoku rules, ensuring that each row, column, and 3x3 subgrid contains unique numbers from 1 to 9. \\
                Additionally, you are provided with a previously explored next state that has been proven to be unsolvable. Use this information to avoid revisiting this failed path and leverage it to make a more informed decision about the current state.\\\\
                
                Current state:\\
                \textcolor{blue}{[[4, 1, 6, 9, 7, 2, 8, 3, 5], [7, 2, 3, 1, 8, 5, 6, 9, 4], [5, 9, 8, 3, 4, 6, 2, 1, 7], [6, 3, 5, 4, 1, 9, 7, 2, 8], [1, 8, 9, 2, 6, 7, 4, 5, 3], [2, 4, 7, 5, 3, 8, 1, 6, 9], [8, 7, 2, 6, 9, 3, 5, 4, 1], [3, 6, 0, 0, 5, 0, 0, 7, 2], [0, 0, 0, 0, 0, 4, 3, 8, 0]]}\\
                Explored next state that leads to an unsolvable path:\\ \textcolor{blue}{[[4, 1, 6, 9, 7, 2, 8, 3, 5], [7, 2, 3, 1, 8, 5, 6, 9, 4], [5, 9, 8, 3, 4, 6, 2, 1, 7], [6, 3, 5, 4, 1, 9, 7, 2, 8], [1, 8, 9, 2, 6, 7, 4, 5, 3], [2, 4, 7, 5, 3, 8, 1, 6, 9], [8, 7, 2, 6, 9, 3, 5, 4, 1], [3, 6, 1, 0, 5, 0, 0, 7, 2], [0, 0, 0, 0, 0, 4, 3, 8, 0]]}\\
                Let's think step by step, considering the failed state to avoid unnecessary exploration. Do not solve using programming.\\
                Choose from (A) Solvable (B) Unsolvable. End your answer with "Answer: (A)" or "Answer: (B)". \\
            \midrule
            \textbf{Sudoku State Transition} \\
            \midrule
            You are given an initial Sudoku puzzle S(0), followed by a sequence of progressive states leading to the current state S(i). Alongside each state, its solvability status L(*) is given. \\
            Your task is to determine the next state by making exactly one move, ensuring progress toward a valid solution. A valid Sudoku solution requires that each row, column, and 3x3 subgrid contains the numbers 1 to 9 without repetition.\\
            
            Additionally, you are provided with a previously explored next state that has been proven to be unsolvable. Use this information to avoid revisiting this failed path.\\

            A move is defined as either:\\
1. Filling: Replacing a 0 in exactly one empty cell with a value from 1 to 9.\\
2. Removing: Replacing a value in exactly one filled cell with 0.\\\\
Initial puzzle: \\
S(0) = \textcolor{blue}{[[0, 1, 0, 0, 7, 0, 8, 3, 0], [0, 2, 0, 0, 0, 0, 6, 0, 4], [5, 9, 0, 0, 0, 0, 0, 1, 0], [0, 0, 5, 0, 1, 9, 0, 0, 0], [1, 0, 0, 2, 6, 0, 0, 5, 3], [2, 4, 7, 0, 0, 8, 0, 0, 9], [0, 7, 0, 6, 9, 0, 0, 0, 1], [3, 0, 0, 0, 5, 0, 0, 7, 2], [0, 0, 0, 0, 0, 4, 3, 8, 0]]}\\
L(0) = \textcolor{blue}{Solvable} \\
Two moves ago:\\
S(i-2) = \textcolor{blue}{[[4, 1, 6, 9, 7, 2, 8, 3, 5], [7, 2, 3, 1, 8, 5, 6, 9, 4], [5, 9, 8, 3, 4, 6, 2, 1, 7], [6, 3, 5, 4, 1, 9, 7, 2, 8], [1, 8, 9, 2, 6, 7, 4, 5, 3], [2, 4, 7, 5, 3, 8, 1, 6, 9], [8, 7, 2, 6, 9, 3, 5, 0, 1], [3, 0, 0, 0, 5, 0, 0, 7, 2], [0, 0, 0, 0, 0, 4, 3, 8, 0]]}\\
L(i-2) = \textcolor{blue}{Solvable} \\
One move ago:\\
S(i-1) = \textcolor{blue}{[[4, 1, 6, 9, 7, 2, 8, 3, 5], [7, 2, 3, 1, 8, 5, 6, 9, 4], [5, 9, 8, 3, 4, 6, 2, 1, 7], [6, 3, 5, 4, 1, 9, 7, 2, 8], [1, 8, 9, 2, 6, 7, 4, 5, 3], [2, 4, 7, 5, 3, 8, 1, 6, 9], [8, 7, 2, 6, 9, 3, 5, 4, 1], [3, 0, 0, 0, 5, 0, 0, 7, 2], [0, 0, 0, 0, 0, 4, 3, 8, 0]]}\\
L(i-1) = \textcolor{blue}{Solvable} \\
Current state:\\
S(i) = \textcolor{blue}{[[4, 1, 6, 9, 7, 2, 8, 3, 5], [7, 2, 3, 1, 8, 5, 6, 9, 4], [5, 9, 8, 3, 4, 6, 2, 1, 7], [6, 3, 5, 4, 1, 9, 7, 2, 8], [1, 8, 9, 2, 6, 7, 4, 5, 3], [2, 4, 7, 5, 3, 8, 1, 6, 9], [8, 7, 2, 6, 9, 3, 5, 4, 1], [3, 6, 0, 0, 5, 0, 0, 7, 2], [0, 0, 0, 0, 0, 4, 3, 8, 0]]}\\
L(i) = \textcolor{blue}{Solvable} \\
Explored next state:\\
S(i+1) = \textcolor{blue}{[[4, 1, 6, 9, 7, 2, 8, 3, 5], [7, 2, 3, 1, 8, 5, 6, 9, 4], [5, 9, 8, 3, 4, 6, 2, 1, 7], [6, 3, 5, 4, 1, 9, 7, 2, 8], [1, 8, 9, 2, 6, 7, 4, 5, 3], [2, 4, 7, 5, 3, 8, 1, 6, 9], [8, 7, 2, 6, 9, 3, 5, 4, 1], [3, 6, 1, 0, 5, 0, 0, 7, 2], [0, 0, 0, 0, 0, 4, 3, 8, 0]]}\\
L(i+1) = \textcolor{blue}{Unsolvable} \\
Let's think step by step. Analyze the progress made so far and determine the immediate next move. 
End your answer with "Next state: \{grid\}", where \{grid\} is in the same python list format as the previous states.\\

            \bottomrule
            \end{tabular}

        }
    }
    \caption{
        Prompt templates for state checking and state transition in Sudoku. \textcolor{blue}{Blue} denotes question specific content.
    }
    \label{tab:prompts-sudoku}
\end{table*}

\begin{table*}[ht]
    \centering
    \resizebox{\linewidth}{!}{
	\setlength{\tabcolsep}{0mm}{
            \begin{tabular}{p{20cm}}
            \toprule
            \textbf{Graph Coloring State Checking} \\
            \midrule
                You are given a graph represented as an adjacency list, where each index corresponds to a vertex, and the list at that index represents its adjacent vertices. You are also given the current coloring state of the graph in a list, where each index represents the corresponding vertex, and the number at that index represents its color (0 indicates an uncolored vertex).\\
                Your task is to determine if this current state can lead to a valid coloring. Specifically, use lookahead techniques to determine if it's possible to color the remaining vertices such that no two adjacent vertices share the same color, using no more than \textcolor{blue}{8} colors in total. \\
                Additionally, you are provided with a previously explored next state that has been proven to be uncolorable. Use this information to avoid revisiting this failed path and leverage it to make a more informed decision about the current state.\\\\

                Graph adjacency list:\\
                \textcolor{blue}{[[1, 4, 6, 7, 10, 11, 12, 14, 15, 16, 17, 18, 19, 21, 23], [0, 3, 5, 10, 12, 14, 15, 16, 17, 21, 22], [7, 10, 11, 12, 17, 18, 19, 21, 23], [1, 6, 7, 10, 11, 12, 14, 18, 21], [0, 6, 7, 10, 11, 12, 15, 20], [1, 6, 7, 8, 10, 11, 20], [0, 3, 4, 5, 9, 10, 11, 12, 14, 15, 16, 19, 20, 21, 22], [0, 2, 3, 4, 5, 10, 11, 13, 14, 18], [5, 9, 11, 13, 14, 16, 17, 19, 22, 23], [6, 8, 10, 14, 17, 23], [0, 1, 2, 3, 4, 5, 6, 7, 9, 13, 14, 16, 19, 22, 23], [0, 2, 3, 4, 5, 6, 7, 8, 12, 13, 14, 17, 19, 20], [0, 1, 2, 3, 4, 6, 11, 14, 15, 16, 18, 19, 21, 23], [7, 8, 10, 11, 14, 15, 18, 20, 22], [0, 1, 3, 6, 7, 8, 9, 10, 11, 12, 13, 15, 16, 18, 19, 20, 21, 23], [0, 1, 4, 6, 12, 13, 14, 17, 18, 19, 20, 21, 23], [0, 1, 6, 8, 10, 12, 14, 17, 21, 22], [0, 1, 2, 8, 9, 11, 15, 16, 18], [0, 2, 3, 7, 12, 13, 14, 15, 17, 19, 21], [0, 2, 6, 8, 10, 11, 12, 14, 15, 18, 23], [4, 5, 6, 11, 13, 14, 15, 21, 22, 23], [0, 1, 2, 3, 6, 12, 14, 15, 16, 18, 20, 23], [1, 6, 8, 10, 13, 16, 20], [0, 2, 8, 9, 10, 12, 14, 15, 19, 20, 21]]}\\
                Current coloring state: \\
                \textcolor{blue}{[1, 2, 1, 1, 2, 1, 3, 3, 2, 1, 4, 4, 5, 1, 6, 4, 7, 3, 0, 0, 0, 0, 0, 0]}\\
                Explored next state that leads to an uncolorable path:\\ \textcolor{blue}{[1, 2, 1, 1, 2, 1, 3, 3, 2, 1, 4, 4, 5, 1, 6, 4, 7, 3, 8, 0, 0, 0, 0, 0]]}\\
                Let's think step by step, considering the failed state to avoid unnecessary exploration. Do not solve using programming.\\
                Choose from (A) Colorable (B) Uncolorable. End your answer with "Answer: (A)" or "Answer: (B)". \\
            \midrule
            \textbf{Graph Coloring State Transition} \\
            \midrule
            You are given a graph represented as an adjacency list, where each index corresponds to a vertex, and the list at that index represents its adjacent vertices. You are also given a sequence of partial coloring states leading to the current coloring state S(i). The coloring state is a list, where each index represents the corresponding vertex in the graph, and the number at that index represents its color (0 indicates an uncolored vertex). Alongside each state, its colorability status L(*) is given. \\
            Your task is to determine the next state by making exactly one move, ensuring progress toward a valid coloring with no more than  \textcolor{blue}{8} colors. A valid coloring requires that no two adjacent vertices share the same color.\\
            Additionally, you are provided with a previously explored next state that has been proven to be uncolorable. Use this information to avoid revisiting this failed path.\\

            A move is defined as either:\\
1. Coloring: Replacing a 0 in exactly one uncolored vertex with a value from 1 to 8.\\
2. Removing a color: Replacing a value in exactly one colored vertex with 0.\\\\
Graph adjacency list: \\
\textcolor{blue}{[[1, 4, 6, 7, 10, 11, 12, 14, 15, 16, 17, 18, 19, 21, 23], [0, 3, 5, 10, 12, 14, 15, 16, 17, 21, 22], [7, 10, 11, 12, 17, 18, 19, 21, 23], [1, 6, 7, 10, 11, 12, 14, 18, 21], [0, 6, 7, 10, 11, 12, 15, 20], [1, 6, 7, 8, 10, 11, 20], [0, 3, 4, 5, 9, 10, 11, 12, 14, 15, 16, 19, 20, 21, 22], [0, 2, 3, 4, 5, 10, 11, 13, 14, 18], [5, 9, 11, 13, 14, 16, 17, 19, 22, 23], [6, 8, 10, 14, 17, 23], [0, 1, 2, 3, 4, 5, 6, 7, 9, 13, 14, 16, 19, 22, 23], [0, 2, 3, 4, 5, 6, 7, 8, 12, 13, 14, 17, 19, 20], [0, 1, 2, 3, 4, 6, 11, 14, 15, 16, 18, 19, 21, 23], [7, 8, 10, 11, 14, 15, 18, 20, 22], [0, 1, 3, 6, 7, 8, 9, 10, 11, 12, 13, 15, 16, 18, 19, 20, 21, 23], [0, 1, 4, 6, 12, 13, 14, 17, 18, 19, 20, 21, 23], [0, 1, 6, 8, 10, 12, 14, 17, 21, 22], [0, 1, 2, 8, 9, 11, 15, 16, 18], [0, 2, 3, 7, 12, 13, 14, 15, 17, 19, 21], [0, 2, 6, 8, 10, 11, 12, 14, 15, 18, 23], [4, 5, 6, 11, 13, 14, 15, 21, 22, 23], [0, 1, 2, 3, 6, 12, 14, 15, 16, 18, 20, 23], [1, 6, 8, 10, 13, 16, 20], [0, 2, 8, 9, 10, 12, 14, 15, 19, 20, 21]]}\\
Two moves ago:\\
S(i-2) = \textcolor{blue}{[1, 2, 1, 1, 2, 1, 3, 3, 2, 1, 4, 4, 5, 1, 6, 4, 0, 0, 0, 0, 0, 0, 0, 0}\\
L(i-2) = \textcolor{blue}{Colorable} \\
One move ago:\\
S(i-1) = \textcolor{blue}{[1, 2, 1, 1, 2, 1, 3, 3, 2, 1, 4, 4, 5, 1, 6, 4, 7, 0, 0, 0, 0, 0, 0, 0]}\\
L(i-1) = \textcolor{blue}{Colorable} \\
Current coloring state:\\
S(i) = \textcolor{blue}{[1, 2, 1, 1, 2, 1, 3, 3, 2, 1, 4, 4, 5, 1, 6, 4, 7, 3, 0, 0, 0, 0, 0, 0]}\\
L(i) = \textcolor{blue}{Colorable} \\
Explored next state:\\
S(i+1) = \textcolor{blue}{[1, 2, 1, 1, 2, 1, 3, 3, 2, 1, 4, 4, 5, 1, 6, 4, 7, 3, 8, 0, 0, 0, 0, 0]}\\
L(i+1) = \textcolor{blue}{Uncolorable} \\
Let's think step by step. Analyze the progress made so far and determine the immediate next move. 
End your answer with "Next state: \{coloring\}", where \{coloring\} is in the same python list format as the previous states.\\

            \bottomrule
            \end{tabular}

        }
    }
    \caption{
        Prompt templates for state checking and state transition in Graph Coloring. \textcolor{blue}{Blue} denotes question specific content.
    }
    \label{tab:prompts-graph}
\end{table*}

\begin{table*}[ht]
    \centering
    \resizebox{\linewidth}{!}{
	\setlength{\tabcolsep}{0mm}{
            \begin{tabular}{p{20cm}}
            \toprule
            \textbf{Game of 24 State Checking} \\
            \midrule
                You are given four numbers and the current calculation state for the Game of 24.\\
                Your task is to determine if this current state can lead to a solvable solution. Specifically, use lookahead techniques to determine if the remaining numbers can be combined using basic arithmetic operations (+ - * /) to reach exactly 24. You must use each number exactly once. \\
                Additionally, you are provided with a previously explored next state that has been proven to be unsolvable. Use this information to avoid revisiting this failed path and leverage it to make a more informed decision about the current state.\\\\

                Numbers:\\
                \textcolor{blue}{[`5', `9', `12', `12']}\\
                Current calculation state: \\
                \textcolor{blue}{[`(5 + 9)', `12', `12']}\\
                Explored next state that leads to an unsolvable path:\\ \textcolor{blue}{[`((5 + 9) + 12)', `12']}\\
                Let's think step by step, considering the failed state to avoid unnecessary exploration. Do not solve using programming.\\
                Choose from (A) Solvable (B) Unsolvable. End your answer with "Answer: (A)" or "Answer: (B)". \\
            \midrule
            \textbf{Game of 24 State Transition} \\
            \midrule
            You are given an initial Game of 24 configuration S(0), followed by a sequence of progressive states leading to the current state S(i). Alongside each state, its solvability status L(*) is given.\\
            Your task is to determine the next state by making exactly one move, ensuring progress toward a valid solution. A valid solution requires using each of the four initial numbers exactly once, using only basic arithmetic operations (+ - * /), and ultimately evaluating to 24. \\
            Additionally, you are provided with a previously explored next state that has been proven to be unsolvable. Use this information to avoid revisiting this failed path.\\

            A move is defined as either:\\
1. Applying an operation: Combining two expressions using a basic arithmetic operation (+ - * /), reducing the number of expressions by 1.\\
2. Reverting an operation: Removing the last operation applied to the expressions, increasing the number of expressions by 1.\\\\
Initial configuration: \\
S(0) = \textcolor{blue}{[`5', `9', `12', `12']}\\
L(0) = \textcolor{blue}{Solvable} \\
Current state:\\
S(i) = \textcolor{blue}{[`(5 + 9)', `12', `12']}\\
L(i) = \textcolor{blue}{Solvable} \\
Explored next state:\\
S(i+1) = \textcolor{blue}{[`((5 + 9) + 12)', `12']}\\
L(i+1) = \textcolor{blue}{Unsolvable} \\
Let's think step by step. Analyze the progress made so far and determine the immediate next move. 
End your answer with "Next state: \{expressions\}", where \{expressions\} is in the same python list format as the previous states.\\

            \bottomrule
            \end{tabular}

        }
    }
    \caption{
        Prompt templates for state checking and state transition in Game of 24. \textcolor{blue}{Blue} denotes question specific content.
    }
    \label{tab:prompts-game24}
\end{table*}

\begin{table*}[ht]
    \centering
    \resizebox{\linewidth}{!}{
	\setlength{\tabcolsep}{0mm}{
            \begin{tabular}{p{20cm}}
            \toprule
            \textbf{Grid Puzzle State Checking} \\
            \midrule
                You are given a partially filled logic grid puzzle represented as a table, where each column corresponds to a specific category, and each row represents attributes of a distinct entry. Empty cells are represented as the empty string (`').\\
                Your task is to determine if this current state can lead to a solvable solution. Specifically, use lookahead techniques to determine if the current configuration can lead to a valid solution under standard logic puzzle constraints (each option in every category must only appear once and adhere to the given clues).\\
                Additionally, you are provided with a previously explored next state that has been proven to be unsolvable. Use this information to avoid revisiting this failed path and leverage it to make a more informed decision about the current state.\\\\

                Question:\\
                \textcolor{blue}{Help Donna make sense of Dr. Finklestein's appointment list for today. Using only the clues below, match the times to the options from names, ailments, and insurers. Remember, as with all grid-based logic puzzles, no option in any category will ever be used more than once.}\\
                Categories:\\
                \textcolor{blue}{times : 9, 10, 11, 12.}\\
\textcolor{blue}{names : Billy, Guy, Paul, Terry.}\\
\textcolor{blue}{ailments : back pain, hip pain, shingles, vertigo.}\\
\textcolor{blue}{insurers : Ambercare, HealthCo, Lifealign, Triflex.}\\
Clues:\\
\textcolor{blue}{1. The person with Lifealign insurance has an appointment sometime before the patient suffering from shingles.} \\
\textcolor{blue}{2. The patient with the 12 noon appointment is either Terry or the patient with Ambercare insurance.}\\
\textcolor{blue}{3. The patient suffering from back pain has an appointment 2 hours before Guy.}\\
\textcolor{blue}{4. The person with Ambercare insurance has an appointment sometime after the person suffering from vertigo.}\\
\textcolor{blue}{5. Neither Billy nor the person suffering from shingles is the person with Lifealign insurance.}\\
\textcolor{blue}{6. The person with the 9:00am appointment is either Paul or the patient suffering from back pain.}\\
\textcolor{blue}{7. The patient with the 10:00am appointment has Triflex insurance.}\\
\textcolor{blue}{8. Of the patient suffering from vertigo and the person with Ambercare insurance, one has the 11:00am appointment and the other is Billy.}\\
                Initial state: \\
                S(0) = \textcolor{blue}{[['times', 'names', 'ailments', 'insurers'], [9, '', '', ''], [10, '', '', ''], [11, '', '', ''], [12, '', '', '']]}\\
                State 1 \textcolor{blue}{(Current state):} \\
                Clue applied: \textcolor{blue}{8} \\
                S(1) = \textcolor{blue}{[['times', 'names', 'ailments', 'insurers'], [9, '', '', ''], [10, '', '', ''], [11, '', 'vertigo', ''], [12, 'Billy', '', 'Ambercare']]}\\
                Explored next state that leads to an unsolvable path:\\ 
                Clue applied: \textcolor{blue}{2} \\
                \textcolor{blue}{S(2) = [['times', 'names', 'ailments', 'insurers'], [9, '', '', ''], [10, 'Terry', '', ''], [11, '', 'vertigo', ''], [12, 'Billy', '', 'Ambercare']]}\\
                Let's think step by step, considering the failed state to avoid unnecessary exploration. Do not solve using programming.\\
                Choose from (A) Solvable (B) Unsolvable. End your answer with "Answer: (A)" or "Answer: (B)". \\
                        \bottomrule
            \end{tabular}

        }
    }
    \caption{
        Prompt template for state checking in Grid Puzzles. \textcolor{blue}{Blue} denotes question specific content.
    }
    \label{tab:prompts-grid-sc}
\end{table*}

\begin{table*}[ht]
    \centering
    \resizebox{\linewidth}{!}{
	\setlength{\tabcolsep}{0mm}{
            \begin{tabular}{p{20cm}}
            \toprule
            \textbf{Grid Puzzle State Transition} \\
            \midrule
            You are given a logic grid puzzle represented as a table, where each column corresponds to a specific category, and each row represents attributes of a distinct entry. Empty cells are represented as the empty string (`'). You are also given a sequence of progressive states from the initial state S(0) to the current state S(n). Alongside each state, its solvability status L(*) is provided.\\
            Your task is to determine the next state by making exactly one move, ensuring progress toward a valid solution. A valid solution requires that each option in every category appears only once, strictly following the given clues. \\
            Additionally, you are provided with a previously explored next state that has been proven to be unsolvable. Use this information to avoid revisiting this failed path.\\

            A move is defined as either:\\
1. Applying a clue: Filling the table with the values indicated by that clue, as long as it does not conflict with any existing clues or placed options.\\
2. Reverting a clue: Removing the last operation applied to the table.\\\\
Question:\\
                \textcolor{blue}{Help Donna make sense of Dr. Finklestein's appointment list for today. Using only the clues below, match the times to the options from names, ailments, and insurers. Remember, as with all grid-based logic puzzles, no option in any category will ever be used more than once.}\\
                Categories:\\
                \textcolor{blue}{times : 9, 10, 11, 12.}\\
\textcolor{blue}{names : Billy, Guy, Paul, Terry.}\\
\textcolor{blue}{ailments : back pain, hip pain, shingles, vertigo.}\\
\textcolor{blue}{insurers : Ambercare, HealthCo, Lifealign, Triflex.}\\
Clues:\\
\textcolor{blue}{1. The person with Lifealign insurance has an appointment sometime before the patient suffering from shingles.} \\
\textcolor{blue}{2. The patient with the 12 noon appointment is either Terry or the patient with Ambercare insurance.}\\
\textcolor{blue}{3. The patient suffering from back pain has an appointment 2 hours before Guy.}\\
\textcolor{blue}{4. The person with Ambercare insurance has an appointment sometime after the person suffering from vertigo.}\\
\textcolor{blue}{5. Neither Billy nor the person suffering from shingles is the person with Lifealign insurance.}\\
\textcolor{blue}{6. The person with the 9:00am appointment is either Paul or the patient suffering from back pain.}\\
\textcolor{blue}{7. The patient with the 10:00am appointment has Triflex insurance.}\\
\textcolor{blue}{8. Of the patient suffering from vertigo and the person with Ambercare insurance, one has the 11:00am appointment and the other is Billy.}\\
Initial state: \\
                S(0) = \textcolor{blue}{[['times', 'names', 'ailments', 'insurers'], [9, '', '', ''], [10, '', '', ''], [11, '', '', ''], [12, '', '', '']]}\\
                L(0) = \textcolor{blue}{Solvable} \\
                State 1 \textcolor{blue}{(Current state):} \\
                Clue applied: \textcolor{blue}{8} \\
                S(1) = \textcolor{blue}{[['times', 'names', 'ailments', 'insurers'], [9, '', '', ''], [10, '', '', ''], [11, '', 'vertigo', ''], [12, 'Billy', '', 'Ambercare']]}\\
                L(1) = \textcolor{blue}{Solvable} \\
                Explored next state:\\ 
                Clue applied: \textcolor{blue}{2} \\
                \textcolor{blue}{S(2) = [['times', 'names', 'ailments', 'insurers'], [9, '', '', ''], [10, 'Terry', '', ''], [11, '', 'vertigo', ''], [12, 'Billy', '', 'Ambercare']]}\\
                \textcolor{blue}{L(2) = Unsolvable} \\
                Let's think step by step. Analyze the progress made so far and determine the immediate next move. End your answer with "Next state: \{table\}", where \{table\} is in the same python list format as the previous states. \\

            \bottomrule
            \end{tabular}

        }
    }
    \caption{
        Prompt template for state transition in Grid Puzzles. \textcolor{blue}{Blue} denotes question specific content.
    }
    \label{tab:prompts-grid-st}
\end{table*}

\begin{table*}[ht]
    \centering
    \resizebox{\linewidth}{!}{
	\setlength{\tabcolsep}{0mm}{
            \begin{tabular}{p{20cm}}
            \toprule
            \textbf{A question from Grid Puzzle State Checking} \\
            \midrule
                \{Grid Puzzle State Checking instructions\}\\
                Categories:\\
                earnings : 2, 5, 8, 11.\\
movies : Amerigo, Easy to Love, Tippecanoe, Zephyr Breeze.\\
directors : Hick Harmon, Micah Moreno, Nan Norman, Odette Olson.\\
Clues:\\
1. Amerigo made 6 million less than the movie directed by Hick Harmon.\\
2. Zephyr Breeze made 6 million more than Amerigo.\\
3. The movie directed by Micah Moreno made somewhat more than Tippecanoe. \\
4. Of the movie directed by Nan Norman and Easy to Love, one made \$11 million and the other made \$2 million. \\
                Initial state: \\
                S(0) = [['earnings', 'movies', 'directors'], [2, '', ''], [5, '', ''], [8, '', ''], [11, '', '']]\\
                State 1 (Current state): \\
                Clue applied: 4 \\
                S(1) = [['earnings', 'movies', 'directors'], [2, '', 'Nan Norman'], [5, '', ''], [8, '', ''], [11, 'Easy to Love', '']]\\
                Explored next state that leads to an unsolvable path:\\ 
                Clue applied: 1 \\
                S(2) = [['earnings', 'movies', 'directors'], [2, '', 'Nan Norman'], [5, 'Amerigo', ''], [8, '', ''], [11, 'Easy to Love', 'Hick Harmon']]\\
\midrule
\textbf{Output from o1 (Misinterpretation of Premises)} \\
\midrule
To determine whether the current state S(1) can lead to a valid solution, we need to analyze the clues carefully while considering the information from the failed state S(2).\\
**Given:**\\
- **Current State S(1):**\\
  - \$2 million: Director Nan Norman\\
  - \$11 million: Movie Easy to Love\\
- **Failed State S(2):**\\
  - Applied Clue 1 leading to an unsolvable path.\\
**Analysis:**\\
1. **Clue 4 Applied in S(1):**\\
   - "Of the movie directed by Nan Norman and Easy to Love, one made \$11 million and the other made \$2 million."\\
   - In S(1), Nan Norman is at \$2 million, and Easy to Love is at \$11 million.\\
2. **Possible Earnings Assignments Based on Clues 1 \& 2:**\\
   - **Clue 1:** "Amerigo made \$6 million less than the movie directed by Hick Harmon."\\
   - **Clue 2:** "Zephyr Breeze made \$6 million more than Amerigo."\\
   - This implies:\\
     - **Option A:**\\
       - Amerigo at \$5 million, Hick Harmon's movie at \$11 million.\\
       - Zephyr Breeze would need to be at \$11 million (which conflicts with Easy to Love at \$11 million).\\
     - **Option B:**\\
       - Amerigo at \$2 million, Hick Harmon's movie at \$8 million.\\
       - Zephyr Breeze would need to be at \$8 million (possible).\\
3. **Conflict with Clue 4:**\\
   - \textcolor{red}{Assigning Amerigo to \$2 million places it with Nan Norman at the same earnings, which violates the rule that each earnings amount is unique per movie and director in this puzzle.}\\
4. **Clue 3 Introduces Further Conflicts:**\\
   - "The movie directed by Micah Moreno made somewhat more than Tippecanoe."\\
   - With limited earnings slots, we cannot assign Micah Moreno and Tippecanoe without conflicts while satisfying previous assignments and clues.\\
5. **Conclusion from Analysis:**\\
   - All possible assignments lead to conflicts or violate the puzzle's constraints.\\
   - The failed state S(2) reinforces that applying Clue 1 after Clue 4 leads to an unsolvable path.\\
   - No alternative assignments in S(1) avoid these conflicts.\\
**Therefore,** considering the current state and the previous failed state, we conclude that the puzzle cannot be solved from S(1).\\
\\
                        \bottomrule
            \end{tabular}

        }
    }
    \caption{
        An example of Misinterpretation of Premises during state checking in Grid Puzzles. The model wrongly concludes that \textcolor{red}{assigning Amerigo to \$2 million alongside Nan Norman violates uniqueness. This is a misinterpretation between the movie and director categories}.
    }
    \label{tab:misintepretation}
\end{table*}

\begin{table*}[ht]
    \centering
    \resizebox{\linewidth}{!}{
	\setlength{\tabcolsep}{0mm}{
            \begin{tabular}{p{20cm}}
            \toprule
            \textbf{A question from Grid Puzzle State Checking} \\
            \midrule
                \{Grid Puzzle State Checking instructions\}\\
                Categories:\\
                dollars: 250, 275, 300, 325, 350, 375.\\
woodworkers : Bonita, Ellen, Josephine, Kendra, Olivia, Patsy.\\
woods : ash, mahogany, oak, poplar, rosewood, walnut.\\
pieces : barstool, bookshelf, coffee table, end table, footstool, hutch.\\
Clues:\\
1. The \$325 piece is the barstool.\\
2. Of the piece made of ash and the barstool, one was crafted by Bonita and the other costs \$250.\\
3. The end table costs 25 dollars more than Ellen's item.\\
4. Ellen's piece costs 50 dollars more than the hutch.\\
5. The bookshelf wasn't made of oak.\\
6. Kendra's item wasn't made of rosewood.\\
7. Neither the coffee table nor the item made of oak is Kendra's piece.\\
8. Josephine's item is either the item made of walnut or the end table.\\
9. Of Bonita's piece and the bookshelf, one was made of poplar and the other costs \$275.\\
10. The piece made of walnut costs 75 dollars more than Patsy's piece.\\
11. Neither the piece made of oak nor the item made of walnut is the coffee table.\\
\{Previous states skipped here... Previous clues applied: 2, 9\}\\
                State 3 (Current state):\\
                Clue applied: 1\\
                S(3) = [['dollars', 'woodworkers', 'woods', 'pieces'], [250, '', 'ash', ''], [275, '', '', 'bookshelf'], [300, '', '', ''], [325, 'Bonita', 'poplar', 'barstool'], [350, '', '', ''], [375, '', '', '']]\\
                Explored next state that leads to an unsolvable path:\\ 
                Clue applied: 3 \\
                S(4) = [['dollars', 'woodworkers', 'woods', 'pieces'], [250, '', 'ash', ''], [275, 'Ellen', '', 'bookshelf'], [300, '', '', 'end table'], [325, 'Bonita', 'poplar', 'barstool'], [350, '', '', ''], [375, '', '', '']]\\
\midrule
\textbf{Clue analysis from o1 (Exploration and Backtracking Issues)} \\
\midrule
To determine if the current state S(3) can lead to a solvable solution, let's see if we can progress without contradictions.\\
1. **Clue 3 \& 4 (Ellen's Item and the End Table):**\\
   - Ellen's item costs \$50 more than the hutch.\\
   - The end table costs \$25 more than Ellen's item.\\
   - Assume Ellen's item is at \$350:\\
     - Hutch would be at \$300 (since \$350 - \$50 = \$300).\\
     - End table would be at \$375 (since \$350 + \$25 = \$375).\\
   - These prices are available, so we assign:\\
     - **\$350 row**: Ellen's item\\
     - **\$375 row**: End table\\
     - **\$300 row**: Hutch\\
2. **Clue 10 (Walnut and Patsy's Item):**\\
   - The piece made of walnut costs \$75 more than Patsy's piece.\\
   - \textcolor{teal}{Potential pairs are (\$300, \$375), (\$275, \$350), or (\$250, \$325).}\\
   - Considering previous assignments:\\
     - \textcolor{red}{Assign walnut to \$375 (end table). Patsy's item would be at \$300 (hutch).}\\
   - Assign:\\
     - **\$375 row**: Walnut, end table\\
     - **\$300 row**: Patsy's item, hutch\\
3. **Clue 8 (Josephine's Item):**\\
   - Josephine's item is either the item made of walnut or the end table.\\
   - Since the end table is at \$375 and made of walnut, Josephine must have the end table.\\
   - Assign:\\
     - **\$375 row**: Josephine, walnut, end table\\
4. **Clue 11 (Coffee Table Not Made of Oak or Walnut):**\\
   - Since walnut is already assigned to the end table, the coffee table can't be made of oak or walnut.\\
   - Remaining woods for the coffee table: Mahogany or Rosewood.\\
5. **Clues 6 \& 7 (Kendra's Item):**\\
   - Kendra's item wasn't made of rosewood and isn't the coffee table or made of oak.\\
   - The only wood left for Kendra that's not rosewood or oak is mahogany.\\
   - Possible pieces are footstool or the piece at \$250 with ash (already assigned to ash).\\
   - There's no available price slot for Kendra's item without causing conflicts.\\
                        \bottomrule
            \end{tabular}
        }
    }
    \caption{
        An example of Exploration and Backtracking Issues during state checking in Grid Puzzles. \textcolor{teal}{The model correctly lists out all possible assignments in step 2,} \textcolor{red}{but did not backtrack and consider other possible assignments when the current assignment leads to conflicts}.
    }
    \label{tab:exploration}
\end{table*}

\begin{table*}[ht]
    \centering
    \resizebox{\linewidth}{!}{
	\setlength{\tabcolsep}{0mm}{
            \begin{tabular}{p{20cm}}
            \toprule
            \textbf{A question from Grid Puzzle State Checking} \\
            \midrule
                \{Grid Puzzle State Checking instructions\}\\
                Categories:\\
                orders : 1, 2, 3, 4, 5, 6.\\
names : Howard, Katrina, Melissa, Nick, Peter, Shari.\\
items : balls, batons, bowling pins, chainsaws, flashlights, mobile phones.\\
towns : Carbon, Dubuque, Menlo, Nice, Quasqueton, Selma.\\
Clues:\\
1. The performer from Quasqueton didn't use balls.\\
2. The person who went sixth wasn't from Carbon.\\
3. The person from Nice didn't use mobile phones.\\
4. Howard went fourth.\\
5. The person who went first didn't use mobile phones.\\
6. The juggler from Menlo was either the juggler who used chainsaws or Nick.\\
7. Of the performer who went sixth and the juggler who went fifth, one used flashlights and the other was from Menlo.\\
8. Melissa was either the juggler who used chainsaws or the performer from Selma.\\
9. Nick performed 3 spots before the juggler from Nice.\\
10. The juggler from Menlo performed 2 spots after Melissa.\\
11. Shari didn't go fifth.\\
Initial state:\\
S(0) = [['orders', 'names', 'items', 'towns'], [1, '', '', ''], [2, '', '', ''], [3, '', '', ''], [4, '', '', ''], [5, '', '', ''], [6, '', '', '']]\\
                State 1 (Current state):\\
                Clue applied: 7\\
                S(1) = [['orders', 'names', 'items', 'towns'], [1, '', '', ''], [2, '', '', ''], [3, '', '', ''], [4, '', '', ''], [5, '', 'flashlights', ''], [6, '', '', 'Menlo']]\\
                Explored next state that leads to an unsolvable path:\\ 
                Clue applied: 10 \\
                S(2) = [['orders', 'names', 'items', 'towns'], [1, '', '', ''], [2, '', '', ''], [3, '', '', ''], [4, 'Melissa', '', ''], [5, '', 'flashlights', ''], [6, '', '', 'Menlo']]\\
\midrule
\textbf{Output from o1 (Inconsistent Reasoning)} \\
\midrule
To determine if the current state S(1) can lead to a solvable solution, let's analyze the clues and the given failed path.\\
**Given Failed Path:** In S(2), assigning "Melissa" to order 4 leads to a contradiction because Clue 4 states "Howard went fourth," so "Melissa" cannot be in order 4. This suggests that assigning "Menlo" to order 6 (as in S(1)) and then applying Clue 10 leads to an unsolvable situation.\\
**Alternative Approach:**\\
1. **Re-evaluate Clue 7:** Clue 7 says, "Of the performer who went sixth and the juggler who went fifth, one used flashlights and the other was from Menlo." In S(1), we assigned "flashlights" to order 5 and "Menlo" to order 6. However, we could instead assign "Menlo" to order 5 and "flashlights" to order 6.\\
2. **Apply Clue 10 with New Assignments:** With "Menlo" at order 5, Clue 10 ("The juggler from Menlo performed 2 spots after Melissa.") means "Melissa" performed at order 3, which is possible since that spot is unassigned.\\
3. **Check for Consistency:** This new assignment doesn't conflict with any other clues. "Howard" is still at order 4 (Clue 4), and "Melissa" is at order 3.\\
**Conclusion:** \textcolor{red}{By adjusting our application of Clue 7 and avoiding the failed path, we find that the puzzle can still be solved from the current state S(1).} The previously explored unsolvable path helps us make a more informed decision to choose an alternative assignment that leads to a solution.\\
                        \bottomrule
            \end{tabular}
        }
    }
    \caption{
        An example of Inconsistent Reasoning during state checking in Grid Puzzles. \textcolor{red}{The model inconsistently suggests changing S(1)'s assignments to solve it, then concludes that the original S(1) is solvable.}
    }
    \label{tab:inconsistent-reasoning}
\end{table*}

\begin{table*}[ht]
    \centering
    \resizebox{\linewidth}{!}{
	\setlength{\tabcolsep}{0mm}{
            \begin{tabular}{p{20cm}}
            \toprule
            \textbf{A question from Grid Puzzle State Checking} \\
            \midrule
                \{Grid Puzzle State Checking instructions\}\\
                Categories:\\
days : 3, 4, 5, 6, 7.\\
requests : clown, petting zoo, photo booth, rock band, superhero.\\
families : Benton, Hayden, Ingram, Stanley, Zimmerman.\\
Clues:\\
1. The party on October 5 is for the Ingrams.\\
2. The booking that requested the clown is sometime after the Ingram family booking.\\
3. The booking on October 4 is either the party that requested the petting zoo or the Hayden family booking.\\
4. The five events are the booking on October 7, the party that requested the rock band, the event that requested the superhero, the Ingram family party and the Benton family party.\\
5. The booking that requested the photo booth is either the event on October 6 or the event on October 3.\\
6. The Ingram family booking is 1 day before the booking that requested the superhero.\\
7. The party that requested the superhero is for the Zimmermans.\\
\{Previous states skipped here... Previous clues applied: 5, 3\}\\
State 3 (Current state):\\
Clue applied: 1\\
S(3) = [['days', 'requests', 'families'], [3, 'photo booth', ''], [4, 'petting zoo', ''], [5, '', 'Ingram'], [6, '', ''], [7, '', 'Hayden']]\\
Explored next state that leads to an unsolvable path:\\
Clue applied: 4\\
S(4) = [['days', 'requests', 'families'], [3, 'photo booth', 'Benton'], [4, 'petting zoo', ''], [5, '', 'Ingram'], [6, 'rock band', ''], [7, '', 'Hayden']]\\
\midrule
\textbf{Clue analysis from o1 (Conflicts Resolving Failure)} \\
\midrule
1. **Clue 7**: The party that requested the superhero is for the Zimmermans.\\
   - We don't have the 'superhero' request assigned yet.\\
2. **Clue 6**: The Ingram family booking is 1 day before the booking that requested the superhero.\\
   - Since Ingram is on October 5, the 'superhero' request must be on October 6.\\
   - From Clue 7, the 'superhero' request is for the Zimmermans.\\
   - So, October 6: Requests 'superhero', Family 'Zimmerman'.\\
3. **Clue 2**: The booking that requested the clown is sometime after the Ingram family booking.\\
   - After October 5, so it must be on October 7. So, October 7: Requests 'clown', Family 'Ingram'.\\
4. **Clue 4**: The five events are:\\
   - The booking on October 7 (we have 'clown' request here).\\
   - The party that requested the rock band (not assigned yet).\\
   - The event that requested the superhero (October 6).\\
   - The Ingram family party (October 5).\\
   - The Benton family party (not assigned yet).\\
   - Since the 'rock band' request and Benton family are among these five events, they must be assigned to the remaining days.\\
5. **Assigning Benton Family:**\\
   - Cannot be on October 5 or 6 (already assigned to Ingram and Zimmerman).\\
   - Cannot be on October 7 (it's the 'clown' request).\\
   - So, Benton family must be on October 3. So, October 3: Requests 'photo booth', Family 'Benton'.\\
6. **Assigning 'rock band' Request:**\\
   - Must be one of the five events.\\
   - Remaining day is October 5 (occupied by Ingram) and October 4.\\
   - October 4 is 'petting zoo' (from Clue 3).\\
   - \textcolor{red}{So, 'rock band' request must be on October 5 with Ingram family. So, October 5: Requests 'rock band', Family 'Ingram'.}\\
7. **Assigning Remaining Family (Hayden and Stanley):**\\
   - From Clue 3, since Hayden is not on October 4, Hayden must be on October 7.\\
   - So, October 7: Requests 'clown', Family 'Hayden'.\\
   - Stanley family is then assigned to October 4.\\
   - So, October 4: Requests 'petting zoo', Family 'Stanley'.\\
**Final Assignment:**\\
- October 3: Requests 'photo booth', Family 'Benton'\\
- October 4: Requests 'petting zoo', Family 'Stanley'\\
- October 5: Requests 'rock band', Family 'Ingram'\\
- October 6: Requests 'superhero', Family 'Zimmerman'\\
- October 7: Requests 'clown', Family 'Hayden'\\
                        \bottomrule
            \end{tabular}
        }
    }
    \caption{
        An example of Conflicts Resolving Failure during state checking in Grid Puzzles. \textcolor{red}{The model assigns 
        'Ingram' and 'rock band' on the same row, which conflicts with clue 4.}
    }
    \label{tab:conflicts}
\end{table*}

\begin{table*}[ht]
    \centering
    \resizebox{\linewidth}{!}{
	\setlength{\tabcolsep}{0mm}{
            \begin{tabular}{p{20cm}}
            \toprule
            \textbf{A question from Grid Puzzle State Checking} \\
            \midrule
                \{Grid Puzzle State Checking instructions\}\\
                Categories:\\
dollars: 250, 275, 300, 325, 350, 375.\\
woodworkers : Bonita, Ellen, Josephine, Kendra, Olivia, Patsy.\\
woods : ash, mahogany, oak, poplar, rosewood, walnut.\\
pieces : barstool, bookshelf, coffee table, end table, footstool, hutch.\\
Clues:\\
1. The \$325 piece is the barstool.\\
2. Of the piece made of ash and the barstool, one was crafted by Bonita and the other costs \$250.\\
3. The end table costs 25 dollars more than Ellen's item.\\
4. Ellen's piece costs 50 dollars more than the hutch.\\
5. The bookshelf wasn't made of oak.\\
6. Kendra's item wasn't made of rosewood.\\
7. Neither the coffee table nor the item made of oak is Kendra's piece.\\
8. Josephine's item is either the item made of walnut or the end table.\\
9. Of Bonita's piece and the bookshelf, one was made of poplar and the other costs \$275.\\
10. The piece made of walnut costs 75 dollars more than Patsy's piece.\\
11. Neither the piece made of oak nor the item made of walnut is the coffee table.\\
\{Previous states skipped here... Previous clues applied: 2, 9, 1, 3, 4, 8\}\\
State 7 (Current state):\\
Clue applied: 10\\
S(7) = [['dollars', 'woodworkers', 'woods', 'pieces'], [250, '', 'ash', ''], [275, 'Patsy', '', 'bookshelf'], [300, '', '', 'hutch'], [325, 'Bonita', 'poplar', 'barstool'], [350, 'Ellen', 'walnut', ''], [375, 'Josephine', '', 'end table']]\\
Explored next state that leads to an unsolvable path:\\
Clue applied: 11\\
S(8) = [['dollars', 'woodworkers', 'woods', 'pieces'], [250, '', 'ash', 'coffee table'], [275, 'Patsy', 'oak', 'bookshelf'], [300, '', '', 'hutch'], [325, 'Bonita', 'poplar', 'barstool'], [350, 'Ellen', 'walnut', 'footstool'], [375, 'Josephine', '', 'end table']]\\
\midrule
\textbf{Clue analysis from o1 (False Premise)} \\
\midrule
1. **Clue 1** is satisfied: The **\$325** piece is the **Barstool**.\\
2. **Clue 2** is satisfied with current assignments: **Ash** piece is at **\$250** (not crafted by **Bonita**). **Barstool** (crafted by **Bonita**) is at **\$325**.\\
3. **Clue 3** is satisfied: **End Table** is at **\$375**. **Ellen's** item is at **\$350**. (\$375 = \$350 + \$25).\\
4. **Clue 4** is satisfied: **Ellen's** piece is at **\$350**. **Hutch** is at **\$300**. (\$350 = \$300 + \$50).\\
5. **Clue 5**: **Bookshelf** (at **\$275**) wasn't made of **Oak**.\\
6. **Clue 6**: **Kendra's** item wasn't made of **Rosewood**.\\
7. **Clue 7**: **Kendra's** item is neither the **Coffee Table** nor made of **Oak**.\\
8. **Clue 8** is satisfied: **Josephine's** item is the **End Table**.\\
9. **Clue 9** is satisfied: **Bonita's** piece (**Barstool**) is at **\$325** and made of **Poplar**. **Bookshelf** is at **\$275**. One is made of **Poplar**, the other costs **\$275**.\\
10. **Clue 10** is satisfied: **Walnut** piece is at **\$350** (**Ellen's** item). **Patsy's** piece is at **\$275**. (\$350 = \$275 + \$75).\\
11. **Clue 11**: **Coffee Table** is neither made of **Oak** nor **Walnut**. \\
**Attempting to Assign Remaining Options:**\\
- \textcolor{red}{**Kendra** cannot be assigned to **\$250** (since that's the **Ash** piece).}\\
- **Kendra** cannot be assigned to **\$275** (already assigned to **Patsy**).\\
- **Kendra** cannot be assigned to **\$300** (piece is **Hutch**, but **Kendra**'s piece isn't specified yet).\\
- **Kendra** cannot be at **\$350** (already assigned to **Ellen**).\\
- **Kendra** cannot be at **\$375** (already assigned to **Josephine**).\\
This leaves no valid position for **Kendra**, given the constraints from the clues.\\
                        \bottomrule
            \end{tabular}
        }
    }
    \caption{
        An example of False Premise during state checking in Grid Puzzles. \textcolor{red}{The question did not mention Kendra cannot be assigned to the Ash piece. This is a nonexistent constraint.}
    }
    \label{tab:false-premise}
\end{table*}

\begin{table*}[ht]
    \centering
    \resizebox{\linewidth}{!}{
	\setlength{\tabcolsep}{0mm}{
            \begin{tabular}{p{20cm}}
            \toprule
            \textbf{A question from Sudoku State Transition} \\
            \midrule
            You are given an initial Sudoku puzzle S(0), followed by a sequence of progressive states leading to the current state S(i). Alongside each state, its solvability status L(*) is given. \\
            Your task is to determine the next state by making exactly one move, ensuring progress toward a valid solution. A valid Sudoku solution requires that each row, column, and 3x3 subgrid contains the numbers 1 to 9 without repetition.\\
            
            Additionally, you are provided with a previously explored next state that has been proven to be unsolvable. Use this information to avoid revisiting this failed path.\\

            A move is defined as either:\\
1. Filling: Replacing a 0 in exactly one empty cell with a value from 1 to 9.\\
2. Removing: Replacing a value in exactly one filled cell with 0.\\\\
Initial puzzle: \\
S(0) = [[5, 3, 0, 0, 2, 7, 0, 0, 0], [0, 6, 7, 0, 0, 0, 0, 0, 1], [4, 8, 0, 3, 0, 0, 0, 0, 9], [0, 0, 0, 8, 0, 5, 4, 0, 0], [8, 0, 2, 0, 9, 1, 0, 3, 0], [3, 1, 0, 4, 0, 2, 0, 7, 0], [9, 0, 0, 6, 0, 0, 0, 4, 5], [0, 0, 0, 0, 1, 4, 0, 0, 7], [0, 0, 0, 0, 8, 0, 2, 6, 0]]\\
L(0) = Solvable \\
Two moves ago:\\
S(i-2) = [[5, 3, 9, 1, 2, 7, 6, 8, 4], [2, 6, 7, 9, 4, 8, 3, 5, 1], [4, 8, 1, 3, 5, 6, 7, 2, 9], [7, 9, 6, 8, 3, 5, 4, 1, 2], [8, 4, 2, 7, 9, 1, 5, 3, 6], [3, 1, 5, 4, 6, 2, 9, 7, 0], [9, 0, 0, 6, 0, 0, 0, 4, 5], [0, 0, 0, 0, 1, 4, 0, 0, 7], [0, 0, 0, 0, 8, 0, 2, 6, 0]]\\
L(i-2) = Solvable \\
One move ago:\\
S(i-1) = [[5, 3, 9, 1, 2, 7, 6, 8, 4], [2, 6, 7, 9, 4, 8, 3, 5, 1], [4, 8, 1, 3, 5, 6, 7, 2, 9], [7, 9, 6, 8, 3, 5, 4, 1, 2], [8, 4, 2, 7, 9, 1, 5, 3, 6], [3, 1, 5, 4, 6, 2, 9, 7, 8], [9, 0, 0, 6, 0, 0, 0, 4, 5], [0, 0, 0, 0, 1, 4, 0, 0, 7], [0, 0, 0, 0, 8, 0, 2, 6, 0]]\\
L(i-1) = Solvable\\
Current state:\\
S(i) = [5, 3, 9, 1, 2, 7, 6, 8, 4], [2, 6, 7, 9, 4, 8, 3, 5, 1], [4, 8, 1, 3, 5, 6, 7, 2, 9], [7, 9, 6, 8, 3, 5, 4, 1, 2], [8, 4, 2, 7, 9, 1, 5, 3, 6], [3, 1, 5, 4, 6, 2, 9, 7, 8], [9, 2, 0, 6, 0, 0, 0, 4, 5], [0, 0, 0, 0, 1, 4, 0, 0, 7], [0, 0, 0, 0, 8, 0, 2, 6, 0]]\\
L(i) = Solvable \\
Explored next state:\\
S(i+1) = [[5, 3, 9, 1, 2, 7, 6, 8, 4], [2, 6, 7, 9, 4, 8, 3, 5, 1], [4, 8, 1, 3, 5, 6, 7, 2, 9], [7, 9, 6, 8, 3, 5, 4, 1, 2], [8, 4, 2, 7, 9, 1, 5, 3, 6], [3, 1, 5, 4, 6, 2, 9, 7, 8], [9, 2, 3, 6, 0, 0, 0, 4, 5], [0, 0, 0, 0, 1, 4, 0, 0, 7], [0, 0, 0, 0, 8, 0, 2, 6, 0]]\\
L(i+1) = Unsolvable \\
Let's think step by step. Analyze the progress made so far and determine the immediate next move. 
End your answer with "Next state: \{grid\}", where \{grid\} is in the same python list format as the previous states.\\
\midrule
\textbf{Output from o1 (Multiple Moves)} \\
\midrule
To determine the next move from the current state S(i), we need to make an informed decision that avoids the previously explored unsolvable path and makes progress toward solving the Sudoku puzzle.\\
**Analysis of Previous Moves:**\\
- **From S(i-2) to S(i-1):** Filled cell at (Row 6, Col 9) with **8**.\\
- **From S(i-1) to S(i):** Filled cell at (Row 7, Col 2) with **2**.\\
- **Explored and found unsolvable S(i+1):** Filled cell at (Row 7, Col 3) with **3**.\\
**Current Empty Cells in S(i) and Potential Moves:**\\
- **Row 7:** Cells at positions (Row 7, Col 3), (Row 7, Col 5), (Row 7, Col 6), (Row 7, Col 7)\\
- **Row 8 and Row 9:** Multiple empty cells\\
We need to choose an empty cell that, when filled, adheres to Sudoku rules.\\
**Choosing Cell at (Row 7, Col 5):**\\
- **Cell Position:** (Row 7, Col 5)\\
- **Current Value:** **0**\\
**Possible Numbers for This Cell:**\\
1. **Numbers Already in Row 7:**\\
- **Existing Numbers:** 9, 2, 6, 4, 5\\
2. **Numbers Already in Column 5:**\\
- **Existing Numbers:** 2, 4, 5, 6, 8, 9, 1, **7 (after previous moves)**\\
3. **Numbers Already in the 3x3 Block:**\\
- **Existing Numbers:** 6, 1, 4, 8, **7 (after previous moves)**\\
**Available Numbers:** The number **3** is not present in the row, column, or block.\\
**Next Move:**\\
- **Action:** Fill cell at (Row 7, Col 5) with **3**.\\
- **Reasoning:** **3** is the only number that doesn't conflict with existing numbers in the row, column, or block.\\
**Next State:**\\
Next state: [[5, 3, 9, 1, 2, 7, 6, 8, 4], [2, 6, 7, 9, 4, 8, 3, 5, 1], [4, 8, 1, 3, 5, 6, 7, 2, 9], [7, 9, 6, 8, 3, 5, 4, 1, 2], [8, 4, 2, 7, 9, 1, 5, 3, 6], [3, 1, 5, 4, 6, 2, 9, 7, 8], [9, 2, 0, 6, \textcolor{red}{7}, \textcolor{red}{3}, 0, 4, 5], [0, 0, 0, 0, 1, 4, 0, 0, 7], [0, 0, 0, 0, 8, 0, 2, 6, 0]]
\\
            \bottomrule
            \end{tabular}

        }
    }
    \caption{
     An example of making multiple moves during state transition in Sudoku. The model fills two numbers, \textcolor{red}{7} and \textcolor{red}{3}, into the current state.
    }
    \label{tab:multiple-moves-example}
\end{table*}

\begin{table*}[ht]
    \centering
    \resizebox{\linewidth}{!}{
	\setlength{\tabcolsep}{0mm}{
            \begin{tabular}{p{20cm}}
            \toprule
            \textbf{A question from Sudoku State Transition} \\
            \midrule
            You are given an initial Sudoku puzzle S(0), followed by a sequence of progressive states leading to the current state S(i). Alongside each state, its solvability status L(*) is given. \\
            Your task is to determine the next state by making exactly one move, ensuring progress toward a valid solution. A valid Sudoku solution requires that each row, column, and 3x3 subgrid contains the numbers 1 to 9 without repetition.\\
            
            Additionally, you are provided with a previously explored next state that has been proven to be unsolvable. Use this information to avoid revisiting this failed path.\\

            A move is defined as either:\\
1. Filling: Replacing a 0 in exactly one empty cell with a value from 1 to 9.\\
2. Removing: Replacing a value in exactly one filled cell with 0.\\\\
Initial puzzle: \\
S(0) = [[0, 0, 0, 3, 6, 1, 0, 9, 0], [5, 0, 9, 0, 0, 8, 0, 0, 0], [0, 1, 8, 0, 0, 0, 0, 2, 0], [8, 0, 4, 6, 0, 0, 3, 0, 0], [7, 3, 0, 2, 0, 9, 0, 0, 1], [0, 0, 0, 0, 7, 0, 6, 8, 0], [0, 0, 2, 0, 0, 0, 0, 6, 4], [6, 0, 5, 1, 4, 0, 0, 0, 8], [0, 9, 0, 0, 5, 0, 2, 0, 7]]\\
L(0) = Solvable \\
Two moves ago:\\
S(i-2) = [[2, 4, 7, 3, 6, 1, 8, 9, 5], [5, 6, 9, 7, 2, 8, 1, 4, 3], [3, 1, 8, 5, 9, 4, 7, 2, 6], [8, 2, 4, 6, 1, 5, 3, 7, 9], [7, 3, 6, 2, 8, 9, 0, 0, 1], [0, 0, 0, 0, 7, 0, 6, 8, 0], [0, 0, 2, 0, 0, 0, 0, 6, 4], [6, 0, 5, 1, 4, 0, 0, 0, 8], [0, 9, 0, 0, 5, 0, 2, 0, 7]]\\
L(i-2) = Solvable \\
One move ago:\\
S(i-1) = [[2, 4, 7, 3, 6, 1, 8, 9, 5], [5, 6, 9, 7, 2, 8, 1, 4, 3], [3, 1, 8, 5, 9, 4, 7, 2, 6], [8, 2, 4, 6, 1, 5, 3, 7, 9], [7, 3, 6, 2, 8, 9, 4, 0, 1], [0, 0, 0, 0, 7, 0, 6, 8, 0], [0, 0, 2, 0, 0, 0, 0, 6, 4], [6, 0, 5, 1, 4, 0, 0, 0, 8], [0, 9, 0, 0, 5, 0, 2, 0, 7]]\\
L(i-1) = Solvable\\
Current state:\\
S(i) = [[2, 4, 7, 3, 6, 1, 8, 9, 5], [5, 6, 9, 7, 2, 8, 1, 4, 3], [3, 1, 8, 5, 9, 4, 7, 2, 6], [8, 2, 4, 6, 1, 5, 3, 7, 9], [7, 3, 6, 2, 8, 9, 4, 5, 1], [0, 0, 0, 0, 7, 0, 6, 8, 0], [0, 0, 2, 0, 0, 0, 0, 6, 4], [6, 0, 5, 1, 4, 0, 0, 0, 8], [0, 9, 0, 0, 5, 0, 2, 0, 7]]\\
L(i) = Solvable \\
Explored next state:\\
S(i+1) = [[2, 4, 7, 3, 6, 1, 8, 9, 5], [5, 6, 9, 7, 2, 8, 1, 4, 3], [3, 1, 8, 5, 9, 4, 7, 2, 6], [8, 2, 4, 6, 1, 5, 3, 7, 9], [7, 3, 6, 2, 8, 9, 4, 5, 1], [1, 0, 0, 0, 7, 0, 6, 8, 0], [0, 0, 2, 0, 0, 0, 0, 6, 4], [6, 0, 5, 1, 4, 0, 0, 0, 8], [0, 9, 0, 0, 5, 0, 2, 0, 7]]\\
L(i+1) = Unsolvable \\
Let's think step by step. Analyze the progress made so far and determine the immediate next move. 
End your answer with "Next state: \{grid\}", where \{grid\} is in the same python list format as the previous states.\\
\midrule
\textbf{Output from o1 (Invalid Move)} \\
\midrule
To solve this problem, we need to find the next valid move from the current Sudoku state S(i), ensuring progress toward a solution and avoiding previously failed paths. Here's the step-by-step analysis:\\
**Step 1: Analyze the Previous Moves**\\
- **From S(i-2) to S(i-1)**: We filled the number **4** at position (row 4, col 6).\\
- **From S(i-1) to S(i)**: We filled the number **5** at position (row 4, col 7).\\
- **Attempted Move in S(i+1)**: Filled the number **1** at position (row 5, col 0), which led to an unsolvable state.\\
**Step 2: Identify Empty Cells in S(i)**\\
List all positions with zeros (empty cells):\\
- **Row 5**: (5,0), (5,1), (5,2), (5,3), (5,5), (5,8)\\
- **Row 6**: (6,0), (6,1), (6,3), (6,4), (6,5), (6,6)\\
- **Row 7**: (7,1), (7,5), (7,6), (7,7)\\
- **Row 8**: (8,0), (8,2), (8,3), (8,5), (8,7)\\
**Step 3: Choose a Cell and Determine Possible Numbers**\\
We select the cell at **(6,1)** for our next move.\\
- **Numbers in Row 6**: 2, 4, 6\\
- **Numbers in Column 1**: 1, 2, 3, 4, 6, 9\\
- **Numbers in the 3x3 Block**: 2, 4, 6, 7, 8\\
- **Possible Numbers for (6,1)**: **5** (since it's not present in the row, column, or block)\\
**Step 4: Make the Move**\\
Fill the number **5** at position **(6,1)** in the current state S(i).\\
**Step 5: Ensure the Move Avoids Previously Failed Paths**\\
We avoid filling **1** at position (5,0) since it led to an unsolvable state. Our chosen move is different and moves us closer to a solution.\\
Next state: [[2, 4, 7, 3, 6, 1, 8, 9, 5], [5, 6, 9, 7, 2, 8, 1, 4, 3], [3, 1, 8, 5, 9, 4, 7, 2, 6], [8, 2, 4, 6, 1, 5, 3, 7, 9], [7, 3, 6, 2, 8, 9, 4, 5, 1], [0, 0, 0, 0, 7, 0, 6, 8, 0], [0, \textcolor{red}{5}, 2, 0, 0, 0, 0, 6, 4], [6, 0, \textcolor{red}{5}, 1, 4, 0, 0, 0, 8], [0, 9, 0, 0, 5, 0, 2, 0, 7]]
\\
            \bottomrule
            \end{tabular}

        }
    }
    \caption{
     An example of making an invalid move during state transition in Sudoku. The model fills \textcolor{red}{5} at position (6,1), which conflicts with \textcolor{red}{5} at position (7,2) as both 5s are in the same 3x3 subgrid.
    }
    \label{tab:invalid-move-example}
\end{table*}

\begin{table*}[ht]
    \centering
    \resizebox{\linewidth}{!}{
	\setlength{\tabcolsep}{0mm}{
            \begin{tabular}{p{20cm}}
            \toprule
            \textbf{A question from Sudoku State Transition} \\
            \midrule
            You are given an initial Sudoku puzzle S(0), followed by a sequence of progressive states leading to the current state S(i). Alongside each state, its solvability status L(*) is given. \\
            Your task is to determine the next state by making exactly one move, ensuring progress toward a valid solution. A valid Sudoku solution requires that each row, column, and 3x3 subgrid contains the numbers 1 to 9 without repetition.\\
            
            Additionally, you are provided with a previously explored next state that has been proven to be unsolvable. Use this information to avoid revisiting this failed path.\\

            A move is defined as either:\\
1. Filling: Replacing a 0 in exactly one empty cell with a value from 1 to 9.\\
2. Removing: Replacing a value in exactly one filled cell with 0.\\\\
Initial puzzle: \\
S(0) = [[0, 0, 6, 4, 0, 2, 0, 8, 0], [3, 7, 0, 0, 5, 0, 0, 9, 0], [2, 0, 8, 9, 0, 7, 4, 0, 0], [0, 0, 0, 0, 0, 9, 0, 0, 0], [0, 1, 0, 8, 0, 6, 0, 4, 3], [0, 0, 5, 7, 1, 0, 0, 0, 0], [0, 0, 0, 0, 0, 4, 0, 2, 7], [0, 8, 9, 2, 0, 0, 3, 0, 6], [0, 0, 2, 3, 0, 0, 8, 1, 0]]\\
L(0) = Solvable \\
Two moves ago:\\
S(i-2) = [[1, 9, 6, 4, 3, 2, 7, 8, 5], [3, 7, 4, 1, 5, 8, 6, 9, 2], [2, 5, 8, 9, 6, 7, 4, 3, 1], [0, 0, 0, 0, 0, 9, 0, 0, 0], [0, 1, 0, 8, 0, 6, 0, 4, 3], [0, 0, 5, 7, 1, 0, 0, 0, 0], [0, 0, 0, 0, 0, 4, 0, 2, 7], [0, 8, 9, 2, 0, 0, 3, 0, 6], [0, 0, 2, 3, 0, 0, 8, 1, 0]]\\
L(i-2) = Solvable \\
One move ago:\\
S(i-1) = [[1, 9, 6, 4, 3, 2, 7, 8, 5], [3, 7, 4, 1, 5, 8, 6, 9, 2], [2, 5, 8, 9, 6, 7, 4, 3, 1], [6, 0, 0, 0, 0, 9, 0, 0, 0], [0, 1, 0, 8, 0, 6, 0, 4, 3], [0, 0, 5, 7, 1, 0, 0, 0, 0], [0, 0, 0, 0, 0, 4, 0, 2, 7], [0, 8, 9, 2, 0, 0, 3, 0, 6], [0, 0, 2, 3, 0, 0, 8, 1, 0]]\\
L(i-1) = Solvable\\
Current state:\\
S(i) = [[1, 9, 6, 4, 3, 2, 7, 8, 5], [3, 7, 4, 1, 5, 8, 6, 9, 2], [2, 5, 8, 9, 6, 7, 4, 3, 1], [6, 2, 0, 0, 0, 9, 0, 0, 0], [0, 1, 0, 8, 0, 6, 0, 4, 3], [0, 0, 5, 7, 1, 0, 0, 0, 0], [0, 0, 0, 0, 0, 4, 0, 2, 7], [0, 8, 9, 2, 0, 0, 3, 0, 6], [0, 0, 2, 3, 0, 0, 8, 1, 0]]\\
L(i) = Solvable \\
Explored next state:\\
S(i+1) = \textcolor{red}{[[1, 9, 6, 4, 3, 2, 7, 8, 5], [3, 7, 4, 1, 5, 8, 6, 9, 2], [2, 5, 8, 9, 6, 7, 4, 3, 1], [6, 2, 7, 0, 0, 9, 0, 0, 0], [0, 1, 0, 8, 0, 6, 0, 4, 3], [0, 0, 5, 7, 1, 0, 0, 0, 0], [0, 0, 0, 0, 0, 4, 0, 2, 7], [0, 8, 9, 2, 0, 0, 3, 0, 6], [0, 0, 2, 3, 0, 0, 8, 1, 0]]}\\
L(i+1) = Unsolvable \\
Let's think step by step. Analyze the progress made so far and determine the immediate next move. 
End your answer with "Next state: \{grid\}", where \{grid\} is in the same python list format as the previous states.\\
\midrule
\textbf{Output from GPT-3.5 (Unsolvable Child)} \\
\midrule
Looking at the current state S(i), we can see that the only empty cell in the third row and third column can be filled with the number 7 to form a valid solution. Therefore, the next move is to fill the cell at S(i)[2][2] with the number 7.\\
Next state: \textcolor{red}{[[1, 9, 6, 4, 3, 2, 7, 8, 5], [3, 7, 4, 1, 5, 8, 6, 9, 2], [2, 5, 8, 9, 6, 7, 4, 3, 1], [6, 2, 7, 0, 0, 9, 0, 0, 0], [0, 1, 0, 8, 0, 6, 0, 4, 3], [0, 0, 5, 7, 1, 0, 0, 0, 0], [0, 0, 0, 0, 0, 4, 0, 2, 7], [0, 8, 9, 2, 0, 0, 3, 0, 6], [0, 0, 2, 3, 0, 0, 8, 1, 0]]}
\\
            \bottomrule
            \end{tabular}

        }
    }
    \caption{
    An example of transitioning to an unsolvable next state during state transition in Sudoku. The model simply fills \textcolor{red}{7}, which is the same move as the unsolvable next state.
    }
    \label{tab:unsolvable-child-example}
\end{table*}

\begin{table*}[ht]
    \centering
    \resizebox{\linewidth}{!}{
	\setlength{\tabcolsep}{0mm}{
            \begin{tabular}{p{20cm}}
            \toprule
            \textbf{A question from Sudoku State Transition} \\
            \midrule
            You are given an initial Sudoku puzzle S(0), followed by a sequence of progressive states leading to the current state S(i). Alongside each state, its solvability status L(*) is given. \\
            Your task is to determine the next state by making exactly one move, ensuring progress toward a valid solution. A valid Sudoku solution requires that each row, column, and 3x3 subgrid contains the numbers 1 to 9 without repetition.\\
            
            Additionally, you are provided with a previously explored next state that has been proven to be unsolvable. Use this information to avoid revisiting this failed path.\\

            A move is defined as either:\\
1. Filling: Replacing a 0 in exactly one empty cell with a value from 1 to 9.\\
2. Removing: Replacing a value in exactly one filled cell with 0.\\\\
Initial puzzle: \\
S(0) = [[0, 0, 6, 4, 0, 2, 0, 8, 0], [3, 7, 0, 0, 5, 0, 0, 9, 0], [2, 0, 8, 9, 0, 7, 4, 0, 0], [0, 0, 0, 0, 0, 9, 0, 0, 0], [0, 1, 0, 8, 0, 6, 0, 4, 3], [0, 0, 5, 7, 1, 0, 0, 0, 0], [0, 0, 0, 0, 0, 4, 0, 2, 7], [0, 8, 9, 2, 0, 0, 3, 0, 6], [0, 0, 2, 3, 0, 0, 8, 1, 0]]\\
L(0) = Solvable \\
Two moves ago:\\
S(i-2) = [[1, 9, 6, 4, 3, 2, 7, 8, 5], [3, 7, 4, 1, 5, 8, 6, 9, 2], [2, 5, 8, 9, 6, 7, 4, 3, 1], [6, 2, 3, 5, 4, 9, 1, 7, 8], [9, 1, 7, 8, 2, 6, 5, 4, 3], [8, 4, 5, 7, 1, 0, 0, 0, 0], [0, 0, 0, 0, 0, 4, 0, 2, 7], [0, 8, 9, 2, 0, 0, 3, 0, 6], [0, 0, 2, 3, 0, 0, 8, 1, 0]]\\
L(i-2) = Solvable \\
One move ago:\\
S(i-1) = [[1, 9, 6, 4, 3, 2, 7, 8, 5], [3, 7, 4, 1, 5, 8, 6, 9, 2], [2, 5, 8, 9, 6, 7, 4, 3, 1], [6, 2, 3, 5, 4, 9, 1, 7, 8], [9, 1, 7, 8, 2, 6, 5, 4, 3], [8, 4, 5, 7, 1, 3, 0, 0, 0], [0, 0, 0, 0, 0, 4, 0, 2, 7], [0, 8, 9, 2, 0, 0, 3, 0, 6], [0, 0, 2, 3, 0, 0, 8, 1, 0]]\\
L(i-1) = Solvable\\
Current state:\\
S(i) = [[1, 9, 6, 4, 3, 2, 7, 8, 5], [3, 7, 4, 1, 5, 8, 6, 9, 2], [2, 5, 8, 9, 6, 7, 4, 3, 1], [6, 2, 3, 5, 4, 9, 1, 7, 8], [9, 1, 7, 8, 2, 6, 5, 4, 3], [8, 4, 5, 7, 1, 3, 9, 0, 0], [0, 0, 0, 0, 0, 4, 0, 2, 7], [0, 8, 9, 2, 0, 0, 3, 0, 6], [0, 0, 2, 3, 0, 0, 8, 1, 0]]\\
L(i) = Unsolvable \\
Explored next state:\\
S(i+1) = [[1, 9, 6, 4, 3, 2, 7, 8, 5], [3, 7, 4, 1, 5, 8, 6, 9, 2], [2, 5, 8, 9, 6, 7, 4, 3, 1], [6, 2, 3, 5, 4, 9, 1, 7, 8], [9, 1, 7, 8, 2, 6, 5, 4, 3], [8, 4, 5, 7, 1, 3, 9, 6, 0], [0, 0, 0, 0, 0, 4, 0, 2, 7], [0, 8, 9, 2, 0, 0, 3, 0, 6], [0, 0, 2, 3, 0, 0, 8, 1, 0]]\\
L(i+1) = Unsolvable \\
Let's think step by step. Analyze the progress made so far and determine the immediate next move. 
End your answer with "Next state: \{grid\}", where \{grid\} is in the same python list format as the previous states.\\
\midrule
\textbf{Output from GPT-3.5 (Backtracking Failure)} \\
\midrule
From the information provided, we can see that the current state is unsolvable, and the next explored state has also been proven to be unsolvable. To progress towards a valid solution, we need to avoid revisiting the explored failed path.\\
Looking at the current state, we see that the empty cell in row 5, column 9 (0-based index) needs to be filled to make progress. We can fill this cell with the number 6 to move towards a valid solution.\\
Therefore, the immediate next move is:\\
Fill: (5, 8) -> 6\\
Next state: [[1, 9, 6, 4, 3, 2, 7, 8, 5], [3, 7, 4, 1, 5, 8, 6, 9, 2], [2, 5, 8, 9, 6, 7, 4, 3, 1], [6, 2, 3, 5, 4, 9, 1, 7, 8], [9, 1, 7, 8, 2, 6, 5, 4, 3], [8, 4, 5, 7, 1, 3, 9, 0, \textcolor{red}{6}], [0, 0, 0, 0, 0, 4, 0, 2, 7], [0, 8, 9, 2, 0, 0, 3, 0, 6], [0, 0, 2, 3, 0, 0, 8, 1, 0]]
\\
            \bottomrule
            \end{tabular}

        }
    }
    \caption{
        An example of failure to return to the parent state during state transition in Sudoku. The model continue to fill \textcolor{red}{6} from an unsolvable state instead of returning to the parent state S(i-1).}
    \label{tab:backtracking-failure-example}
\end{table*}

\begin{table*}[ht]
    \centering
    \resizebox{\linewidth}{!}{
	\setlength{\tabcolsep}{0mm}{
            \begin{tabular}{p{20cm}}
            \toprule
            \textbf{A question from Sudoku State Transition} \\
            \midrule
            You are given an initial Sudoku puzzle S(0), followed by a sequence of progressive states leading to the current state S(i). Alongside each state, its solvability status L(*) is given. \\
            Your task is to determine the next state by making exactly one move, ensuring progress toward a valid solution. A valid Sudoku solution requires that each row, column, and 3x3 subgrid contains the numbers 1 to 9 without repetition.\\
            
            Additionally, you are provided with a previously explored next state that has been proven to be unsolvable. Use this information to avoid revisiting this failed path.\\

            A move is defined as either:\\
1. Filling: Replacing a 0 in exactly one empty cell with a value from 1 to 9.\\
2. Removing: Replacing a value in exactly one filled cell with 0.\\\\
Initial puzzle: \\
S(0) = [[0, 3, 5, 0, 0, 0, 8, 0, 1], [0, 0, 0, 0, 4, 0, 0, 0, 0], [2, 7, 0, 9, 0, 0, 6, 0, 3], [0, 8, 0, 3, 1, 0, 9, 0, 4], [0, 0, 9, 5, 2, 7, 0, 0, 0], [0, 1, 7, 0, 0, 9, 0, 6, 0], [0, 0, 8, 7, 0, 0, 2, 5, 0], [9, 0, 0, 0, 0, 5, 0, 1, 0], [0, 0, 4, 2, 0, 0, 0, 8, 6]]\\
L(0) = Solvable \\
Two moves ago:\\
S(i-2) = [[4, 3, 5, 6, 7, 2, 8, 9, 1], [8, 9, 6, 1, 4, 3, 7, 2, 5], [2, 7, 1, 9, 5, 8, 6, 4, 3], [5, 8, 2, 3, 1, 6, 9, 7, 4], [6, 4, 9, 5, 2, 7, 1, 3, 8], [3, 1, 7, 4, 8, 9, 5, 6, 2], [1, 0, 8, 7, 0, 0, 2, 5, 0], [9, 0, 0, 0, 0, 5, 0, 1, 0], [0, 0, 4, 2, 0, 0, 0, 8, 6]]\\
L(i-2) = Solvable \\
One move ago:\\
S(i-1) = [[4, 3, 5, 6, 7, 2, 8, 9, 1], [8, 9, 6, 1, 4, 3, 7, 2, 5], [2, 7, 1, 9, 5, 8, 6, 4, 3], [5, 8, 2, 3, 1, 6, 9, 7, 4], [6, 4, 9, 5, 2, 7, 1, 3, 8], [3, 1, 7, 4, 8, 9, 5, 6, 2], [1, 6, 8, 7, 0, 0, 2, 5, 0], [9, 0, 0, 0, 0, 5, 0, 1, 0], [0, 0, 4, 2, 0, 0, 0, 8, 6]]\\
L(i-1) = Solvable\\
Current state:\\
S(i) = [[4, 3, 5, 6, 7, 2, 8, 9, 1], [8, 9, 6, 1, 4, 3, 7, 2, 5], [2, 7, 1, 9, 5, 8, 6, 4, 3], [5, 8, 2, 3, 1, 6, 9, 7, 4], [6, 4, 9, 5, 2, 7, 1, 3, 8], [3, 1, 7, 4, 8, 9, 5, 6, 2], [1, 6, 8, 7, 9, 0, 2, 5, 0], [9, 0, 0, 0, 0, 5, 0, 1, 0], [0, 0, 4, 2, 0, 0, 0, 8, 6]]\\
L(i) = Unsolvable \\
Explored next state:\\
S(i+1) = [[4, 3, 5, 6, 7, 2, 8, 9, 1], [8, 9, 6, 1, 4, 3, 7, 2, 5], [2, 7, 1, 9, 5, 8, 6, 4, 3], [5, 8, 2, 3, 1, 6, 9, 7, 4], [6, 4, 9, 5, 2, 7, 1, 3, 8], [3, 1, 7, 4, 8, 9, 5, 6, 2], [1, 6, 8, 7, 9, 4, 2, 5, 0], [9, 0, 0, 0, 0, 5, 0, 1, 0], [0, 0, 4, 2, 0, 0, 0, 8, 6]]\\
L(i+1) = Unsolvable \\
Let's think step by step. Analyze the progress made so far and determine the immediate next move. 
End your answer with "Next state: \{grid\}", where \{grid\} is in the same python list format as the previous states.\\
\midrule
\textbf{Output from o1 (Sibling)} \\
\midrule
To determine the next move, let's analyze the progress made so far:\\
1. **From S(0) to S(i-2):** The first six rows of the puzzle were filled, making significant progress.\\
2. **From S(i-2) to S(i-1):** The number `6` was filled in row 6, column 1, advancing the puzzle.\\
3. **From S(i-1) to S(i):** The number `9` was filled in row 6, column 4. However, this led to an unsolvable state.\\
4. **Explored S(i+1):** Filling `4` in row 6, column 5 also resulted in an unsolvable state.\\
**Conclusion:** Filling `9` at position [6][4] causes a conflict, making the puzzle unsolvable. To avoid this path and progress towards a solution, we should consider an alternative number for that cell.\\
**Analyzing possible numbers for position [6][4]:**\\
- **Numbers in the same row (row 6):** 1, 6, 7, 8, 2, 5\\
- **Numbers in the same column (column 4):** 7, 4, 5, 1, 2, 8\\
- **Numbers in the same 3x3 block:** 7, 2, 5\\
**Possible numbers for [6][4]:** Numbers from 1 to 9 not in the above lists: **3** and **9**\\
Since `9` leads to an unsolvable state, we should try **`3`** at position [6][4].\\
**Next move:** Fill `3` in row 6, column 4.\\
Next state: [[4, 3, 5, 6, 7, 2, 8, 9, 1], [8, 9, 6, 1, 4, 3, 7, 2, 5], [2, 7, 1, 9, 5, 8, 6, 4, 3], [5, 8, 2, 3, 1, 6, 9, 7, 4], [6, 4, 9, 5, 2, 7, 1, 3, 8], [3, 1, 7, 4, 8, 9, 5, 6, 2], [1, 6, 8, 7, \textcolor{red}{3}, 0, 2, 5, 0], [9, 0, 0, 0, 0, 5, 0, 1, 0], [0, 0, 4, 2, 0, 0, 0, 8, 6]]
\\
            \bottomrule
            \end{tabular}

        }
    }
    \caption{
        An example of making an additional move to a sibling state after backtracking during state transition in Sudoku. The model correctly backtracks by removing 9, but makes an additional move to fill \textcolor{red}{3} at position (6,4). This violates the instruction of making only one move each step.
    }
    \label{tab:sibling-example}
\end{table*}

\begin{table*}[!t]
\small
\centering
\resizebox{1.0\linewidth}{!}{
\setlength{\tabcolsep}{0mm}{
\begin{tabular}{p{4cm}@{~~} @{~}p{7cm}@{~~} @{~}p{7cm}}
\toprule
\textbf{Question} & \textbf{Model 1 (w/o Mixed Training)} & \textbf{Model 2 (w/ Mixed Training)}\\
\midrule
In Pascal's Triangle, the third row has four elements, and the sum of these elements is 8. What is the sum of the elements in the tenth row? (Note that we are counting the top row with a single 1 as the 0th row). & To determine the sum of the elements in the tenth row of Pascal's Triangle, we use the property that the sum of the elements in the nth row is $2^n$. Since the top row is counted as the 0th row, \textcolor{red}{the tenth row corresponds to the 9th row. Calculating $2^9$, we get 512.}
& To determine the sum of the elements in the tenth row of Pascal's Triangle, we use the property that the sum of the elements in the nth row is $2^n$. \textcolor{teal}{Given that the third row has a sum of 8, we can verify this by calculating $2^3 = 8$, which matches the given information.} Therefore, the sum of the elements in the tenth row is $2^{10} = 1024$.\\
\midrule

Determine the number of ways to arrange the letters of the word TEPEE. & Step 1: Identify the letters and their frequencies. The word TEPEE consists of the following letters: - T: 1 occurrence \textcolor{red}{- E: 2 occurrences - P: 2 occurrences.} Step 2: Apply the formula for permutations of a multiset. The formula for the number of distinct permutations of a word with repeated letters is: $\text{Number of permutations} = \frac{n!}{n_1! \times n_2! \times\dots \times n_k!}$ where: $n$ is the total number of letters, $n_1, n_2, \dots, n_k$ are the frequencies of the repeated letters. Step 3: Plug in the values. For the word TEPEE: $n = 5$ (since there are 5 letters). $n_1 = 1$ (for T). $n_2 = 2$ (for E). $n_3 = 2$ (for P). $\text{Number of permutations} = \frac{5!}{1!\times 2!\times 2!} = \frac{120}{1 \times 2 \times 2} = \frac{120}{4} = 30$.
& 1. Identify the Letters and Their Frequencies: The word TEPEE consists of 5 letters. The letters are: T, E, P, E, E. Frequencies: - T: 1 - E: 3 - P: 1 \textcolor{red}{- C: 1} \textcolor{teal}{ Note: The original problem didn't mention `C', so we assume it's not part of the word.} 2. Calculate the Total Number of Arrangements: The formula for permutations of a multiset is: $\text{Number of arrangements} = \frac{n!}{n_1! \times n_2! \times \dots \times n_k!}$ where $n$ is the total number of items, and $n_1, n_2, \dots, n_k$ are the frequencies of each distinct item. Applying this to TEPEE: $\text{Number of arrangements} = \frac{5!}{3! \times 1! \times 1!} = \frac{120}{6 \times 1 \times 1} = \frac{120}{6} = 20$ \\
\bottomrule
\end{tabular}
}
}
\caption{Two examples of how the 1.5B model shows more evidence of verification and self-correction in its thought process after mixed training. \textcolor{red}{Red} denotes where the error starts to occur. \textcolor{teal}{Green} denotes signs of verification or self-correction.}
\label{tab:math-quality-analysis} 
\end{table*}

\end{document}